%% file: main.tex
\documentclass[10pt,twocolumn,letterpaper]{article}

\usepackage[algorithms]{wacv}      


\definecolor{wacvblue}{rgb}{0.21,0.49,0.74}
\usepackage[pagebackref,breaklinks,colorlinks,allcolors=wacvblue]{hyperref}
\usepackage[table]{xcolor} 
\usepackage{float}
\usepackage{multirow}
\usepackage{stfloats}
\def\wacvPaperID{1205} 
\def\confName{WACV}
\def\confYear{2026}

\title{IMKD: Intensity-Aware Multi-Level Knowledge Distillation for Camera-Radar Fusion}

\author{
Shashank Mishra$^{1}$ \quad
Karan Patil$^{1,2}$ \quad
Didier Stricker$^{1,2}$ \quad
Jason Rambach$^{1}$\\[4pt]
$^{1}$German Research Center for Artificial Intelligence (DFKI) 
$^{2}$RPTU\\[4pt]
}

\begin{document}
\maketitle
\input{sec/0_abstract}    
\input{sec/1_intro}
\input{sec/2_realted_work}
\input{sec/3_methodology}
\input{sec/4_experiments}
\input{sec/5_conclusion}
\input{sec/acknowledgement}
{
    \small
    \bibliographystyle{ieeenat_fullname}
    \bibliography{main}
}
\input{sec/6_supplementary.tex}

\end{document}

%% file: sec/0_abstract.tex
\begin{abstract}
High-performance Radar-Camera 3D object detection can be achieved by leveraging knowledge distillation without using LiDAR at inference time. However, existing distillation methods typically transfer modality-specific features directly to each sensor, which can distort their unique characteristics and degrade their individual strengths. To address this, we introduce IMKD, a radar-camera fusion framework based on multi-level knowledge distillation that preserves each sensor’s intrinsic characteristics while amplifying their complementary strengths. IMKD applies a three-stage, intensity-aware distillation strategy to enrich the fused representation across the architecture: (1) LiDAR-to-Radar intensity-aware feature distillation to enhance radar representations with fine-grained structural cues, (2) LiDAR-to-Fused feature intensity-guided distillation to selectively highlight useful geometry and depth information at the fusion level, fostering complementarity between the modalities rather than forcing them to align, and (3) Camera-Radar intensity-guided fusion mechanism that facilitates effective feature alignment and calibration. Extensive experiments on the nuScenes benchmark show that IMKD reaches 67.0\% NDS and 61.0\% mAP, outperforming all prior distillation-based radar-camera fusion methods. Our code and models are available at: \url{https://github.com/dfki-av/IMKD/}.
\end{abstract}

%% file: sec/1_intro.tex
\section{Introduction}
\label{sec:intro}

Bird’s Eye View (BEV) has become the dominant representation for 3D perception in autonomous systems due to its structured spatial layout and planning compatibility. BEV maps are constructed using LiDAR, cameras, and radar—each with distinct characteristics. LiDAR offers precise depth and structure, making it highly effective for 3D detection \cite{mao2021voxel, yin2021center, shi2023pv, hu2022afdetv2}, but its high cost and limited range hinder adoption. Cameras provide rich texture but struggle with depth and low light. Radar is robust in adverse weather and long-range detection but suffers from low spatial resolution and noise.

To leverage cost-effective setups, recent works have explored knowledge distillation (KD) to transfer information from LiDAR to camera or radar-based models \cite{zhao2024crkd, kim2024labeldistill, klingner2023x3kd}. However, most existing approaches distill knowledge independently to each modality, often forcing radar or camera features to mimic LiDAR representations. This direct one-to-one transfer overlooks the unique characteristics of each sensor and can introduce representational conflicts, limiting the effectiveness of the distillation process. Furthermore, the intermediate representations chosen for distillation are often suboptimal, missing opportunities to enhance cross-modal fusion through better guidance signals.

\begin{figure}[t]
  \centering
    \includegraphics[width=0.9\linewidth]{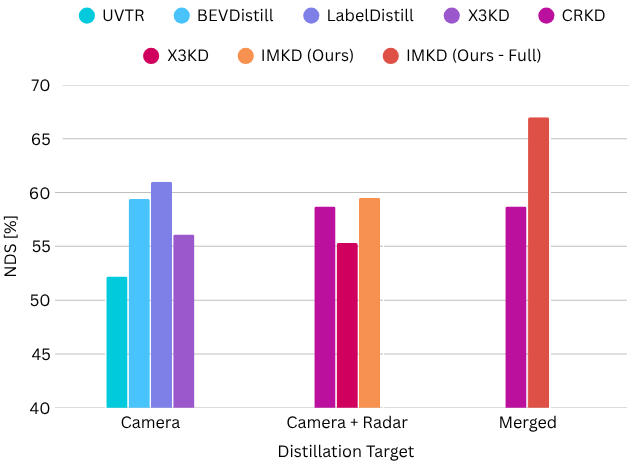}
   \caption{Comparison of KD methods grouped by distillation target. IMKD’s intensity-based knowledge distillation achieves the highest performance.}
   \label{fig:imkd_comparison}
\end{figure}

In this paper, we introduce IMKD, an Intensity-Aware Multi-Level Knowledge Distillation framework that enhances camera–radar representations through cross-modal supervision. Prior methods have explored distillation either at the modality level or in the fused space, but often without accounting for sensor-specific reliability. IMKD addresses this by supervising the fused camera–radar BEV representation with LiDAR as a privileged modality, where intensity serves as a reliability prior that highlights geometrically consistent regions. This enables structural and depth-rich cues to be transferred more effectively into the fused representation while preserving radar’s robustness, leading to improved alignment, stability, and confidence in BEV features.


The intensity-guided distillation mechanism modulates supervision based on LiDAR confidence, emphasizing informative regions while down-weighting ambiguous signals. This adaptive weighting prevents overfitting to modality inconsistencies and, when applied across early, middle, and late fusion stages, yields stable feature alignment and consistent cross-modal refinement.



Beyond modality-specific distillation, we shift focus to the fused representation itself. Performing knowledge distillation in this joint feature space allows supervision to act where cross-modal interactions are already encoded. This leads to better spatial reasoning, stronger synergy between modalities, and ultimately improved detection performance.

Finally, we generalize the use of intensity-aware supervision beyond LiDAR by introducing an intensity-guided radar-camera fusion module. This module estimates confidence from both sensors to guide feature fusion. To further improve the fused representation, we incorporate structured supervision from ground truth labels, offering a reliable signal that remains robust to sensor noise and occlusions. Together, these additions strengthen cross-modal learning and reduce dependence on LiDAR-specific guidance.

To validate our approach, we conduct extensive experiments on the nuScenes dataset \cite{caesar2020nuscenes}. IMKD outperforms prior camera-radar knowledge distillation methods and establishes a new benchmark for effective cross-modal supervision in cost-efficient 3D perception.

The main contributions of this paper are listed as follows:
\begin{itemize}
    \item We present IMKD, an Intensity-Aware Multi-Level Knowledge Distillation framework that enhances camera–radar fusion for 3D object detection, achieving state-of-the-art results among KD-based methods on the nuScenes benchmark \cite{caesar2020nuscenes}.
    \item We design an intensity-aware distillation strategy that preserves the strengths of each sensor modality by guiding knowledge transfer based on high-confidence LiDAR cues. This is applied at multiple stages of the pipeline, enhancing both radar and fused features.
    \item We perform knowledge distillation in the joint fused feature space instead of individual modalities, allowing supervision to operate where cross-modal cues are already integrated, leading to better spatial reasoning and more robust predictions.
    \item  We introduce an intensity-aware radar-camera fusion module that improves fusion using sensor confidence cues.
\end{itemize}

%% file: sec/2_realted_work.tex
\section{Related Work}
\label{sec:related_work}
\subsection{3D Object Detection with Multi-Sensor Fusion}
Multi-modal 3D object detection combines complementary sensors to boost perception performance. LiDAR-Camera (LC) fusion remains the most accurate setup across datasets \cite{caesar2020nuscenes, mei2022waymo, geiger2012we}, implemented via early fusion \cite{sindagi2019mvx, vora2020pointpainting, wang2021pointaugmenting}, feature-level fusion \cite{li2022voxel, li2022unifying, yoo20203d}, and BEV-based methods \cite{liang2022bevfusion, liu2023bevfusion}. However, LiDAR’s high cost limits scalability, positioning Camera-Radar (CR) fusion as a cost-effective alternative for long-range, all-weather perception.

CR fusion is more challenging due to view misalignment and sparse radar signals. Early methods like CenterFusion \cite{nabati2021centerfusion} and RadarNet \cite{yang2020radarnet} established radar-image associations and multi-level fusion. Recent BEV-based models—MVFusion \cite{wu2023mvfusion}, RADIANT \cite{long2023radiant}, CRAFT \cite{kim2023craft}, CRN \cite{kim2023crn}, and RCFusion \cite{zheng2023rcfusion}—focus on improving cross-modal alignment and radar feature aggregation. RCBEVDet \cite{lin2024rcbevdet} and RCBEVDet++ \cite{lin2024rcbevdet++} further refine this pipeline with enhanced fusion strategies.

Building on these advances, our work incorporates knowledge distillation to transfer LiDAR’s geometric cues into the CR pipeline, improving fused feature quality while maintaining efficiency.

\subsection{Cross-modality Knowledge Distillation}
In 3D object detection, traditional Knowledge Distillation (KD) approaches often maintain the same modality for teacher and student models, such as LiDAR-to-LiDAR (L2L) \cite{wang2021object, wei2022lidar, yang2022towards, zhang2023pointdistiller} or Camera-to-Camera (C2C) \cite{li2023bev, zeng2023distilling, zhang2022structured}. However, cross-modality KD, which distills knowledge between different sensor modalities, has gained increasing attention. Established cross-modality KD paradigms include LiDAR-to-Camera (L2C) \cite{chong2022monodistill, chen2022bevdistill, guo2021liga, hong2022cross, li2022unifying, liu2023stereodistill} and Camera-to-LiDAR (C2L) \cite{sautier2022image, tang2024sparseocc}, with recent advancements exploring fusion-based KD. Methods such as UniDistill \cite{zhou2023unidistill} and DistillBEV \cite{wang2023distillbev} unify features into a shared BEV space to facilitate L2C and LC-to-Camera (LC2C) distillation. Additionally, LabelDistill \cite{kim2024labeldistill} demonstrates effective label-based distillation for camera and LiDAR models, ensuring robust supervision without relying solely on feature-space alignment. X3KD \cite{klingner2023x3kd} and CRKD \cite{zhao2024crkd} extend KD to LiDAR-Camera-to-Camera-Radar (LC2CR) by introducing adaptive feature alignment, enabling radar-aware distillation while mitigating domain discrepancies. 

While CRKD \cite{zhao2024crkd} explored fused-to-fused distillation, prior works did not explicitly model modality-to-merged transfer with reliability-aware guidance. IMKD addresses this gap by leveraging LiDAR and label supervision to enhance camera–radar fusion, introducing spatial, depth, and structural cues through intensity-guided distillation.

%% file: sec/3_methodology.tex
\section{Intensity-Aware Multi-Level Knowledge Distillation Framework}
\label{sec:methodology}

\begin{figure*}[t]
  \centering
    \includegraphics[width=1\linewidth]{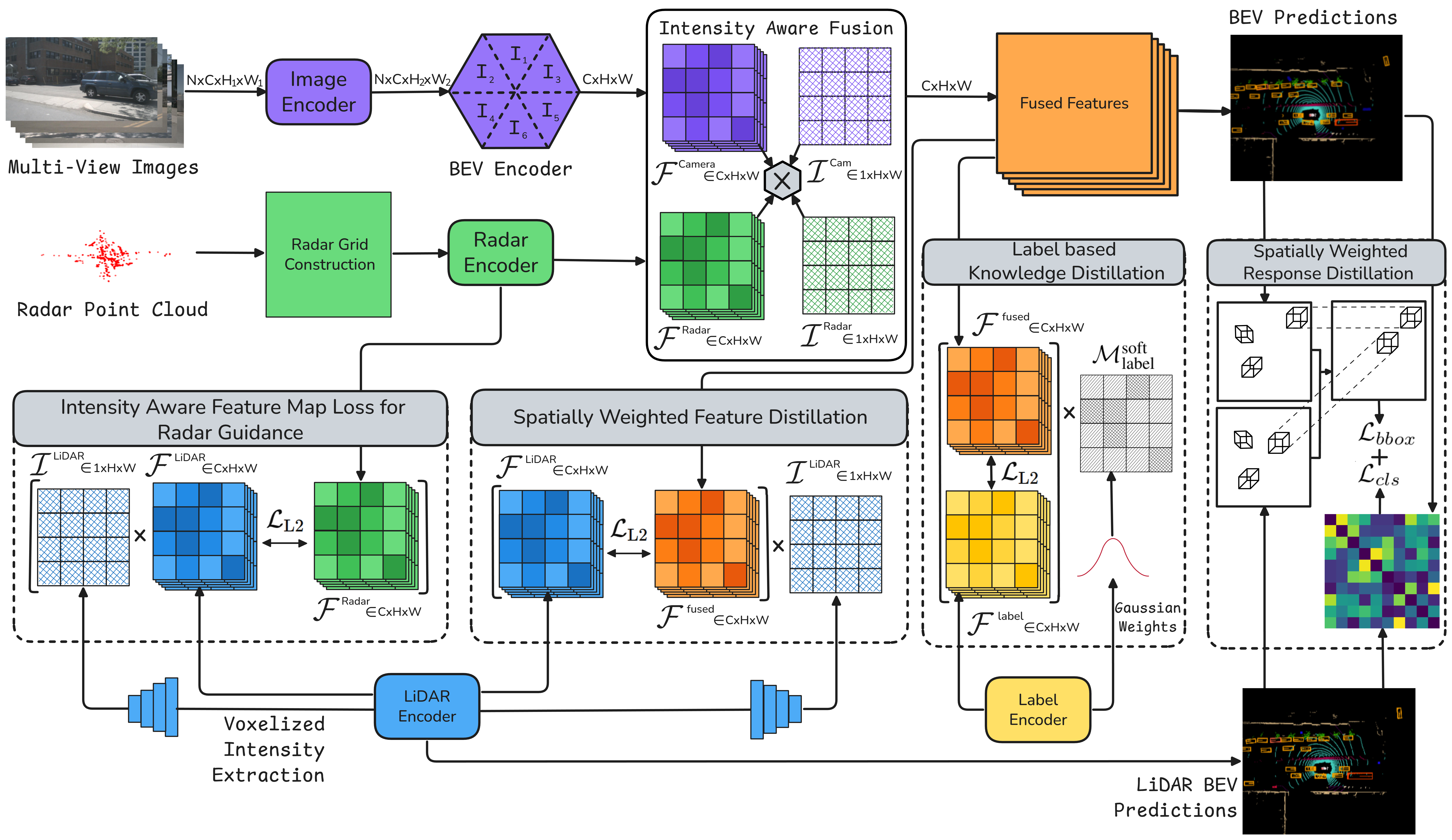}
   \caption{Overview of the proposed Intensity-Aware Multi-Level Knowledge Distillation.}
   \label{fig:imkd_arc}
\end{figure*}

IMKD is a framework designed to overcome the shortcomings of existing knowledge distillation approaches for sensor fusion by enhancing radar representation and introducing intensity-aware, multi-level supervision.

Multi-view image features are lifted to BEV using radar-guided depth, while radar features are encoded via a learnable radar-to-grid module. Fusion is performed using intensity-aware deformable cross-attention, leveraging modality-specific intensity maps for precise alignment. LiDAR-derived features refine radar representations before distillation. LiDAR distillation injects spatial priors, while label distillation provides clean, uncertainty-free supervision. At inference, only camera and radar branches are used for efficient deployment.

\subsection{Camera Feature Extraction}
We extract features from $N$ multi-view images ${I_{1}, \cdots, I{_N}}$ using a convolutional backbone, producing downsampled feature maps $\mathcal{F}_{\mathcal{I}}$ at a resolution of $1/16$ for each view. These features are refined through additional convolutional layers to generate a context-rich perspective-view feature map $\mathbf{C}_{\mathcal{I}}^{\mathcal{PV}} \in \mathbb{R}^{N \times C \times H \times W}$:
\begin{align}
\label{equation1}
        \mathbf{C}_{\mathcal{I}}^{\mathcal{PV}} = Conv(\mathcal{F_{\mathcal{I}}}) \notag \\
        \mathcal{D}_\mathcal{I}(u,v) = \text{Softmax}(Conv(\mathcal{F}_{\mathcal{I}})(u,v)),     
\end{align}
where $(u,v)$ denotes pixel coordinates in the image plane and $\mathcal{D}_\mathcal{I} \in \mathbb{R}^{N \times D \times H \times W}$ is the predicted per-pixel depth distribution across $D$ discrete bins. Following the depth-guided view transformation approach \cite{philion2020lift}, we lift the perspective-view features into a frustum-aligned 3D representation $\mathbf{C}_{\mathcal{I}}^{\mathcal{FV}} \in \mathbb{R}^{N \times C \times D \times H \times W}$:
\begin{equation}
\label{equation2}
\mathbf{C}_{\mathcal{I}}^{\mathcal{FV}} = \text{Conv}(\mathbf{C}_{\mathcal{I}}^{\mathcal{PV}} \otimes \mathcal{D}_\mathcal{I}),
\end{equation}
where $\otimes$ denotes the outer product between the feature maps and the depth probabilities. The resulting frustum-view features are later aggregated across views and projected into the BEV space as $\mathcal{F}^{\text{Camera}}$.
\subsection{Radar Feature Extraction}
Our radar processing pipeline transforms sparse, noisy point cloud measurements into dense BEV features suitable for fusion. To maximize the representational power of radar while preserving its sparsity, we design a learnable radar grid construction module followed by a compact yet effective radar encoder.
\subsubsection{Radar Grid Construction}
Let $\mathcal{P}_{\text{Radar}} = {(x_i, y_i, z_i, v_i, \text{RCS}_i)}_{i \in \{1, \ldots, M\}}$ be the set of $M$ raw radar detections per frame, where $(x_i, y_i, z_i)$ is the 3D position, $v_i$ is the compensated Doppler velocity, and $\text{RCS}_i$ is the radar cross-section. Each point is first embedded using a learnable MLP $\phi$:

\begin{equation}
\label{equation3}
\mathbf{f}_i = \phi(x_i, y_i, z_i, v_i, \text{RCS}_i) \in \mathbb{R}^C
\end{equation}

We then project each radar point onto the BEV plane and map it to a 2D grid cell $(u_i, v_i)$. Instead of using handcrafted statistics, we aggregate features per cell using a differentiable, channel-wise max pooling across all points within that cell:
\begin{align}
\label{equation4}
\mathbf{G}_{\mathcal{R}}(u, v) = \text{MaxPool} \left( { \mathbf{f}_i \mid (u_i, v_i) = (u, v) } \right), \quad \\\mathbf{G}_{\mathcal{R}} \in \mathbb{R}^{C \times H \times W} \notag
\end{align}

This learnable radar-to-grid mapping enables the model to adaptively encode semantic and spatial patterns from raw radar points, replacing brittle hand-engineered descriptors.

\subsubsection{Radar Encoder}
To extract higher-level features from the radar grid, we design a lightweight encoder $\mathcal{E}_{\mathcal{R}}$ that adapts sparse convolutional designs and point-based reasoning. It consists of 16 convolutional layers grouped into residual blocks with BatchNorm and ReLU activations. Two downsampling stages progressively increase channel depth while reducing spatial resolution, producing BEV features:

\begin{equation}
\label{equation5}
\mathcal{F}^{\text{Radar}} = \mathcal{E}_{\mathcal{R}}(\mathbf{G}_{\mathcal{R}}) \in \mathbb{R}^{C \times H \times W}
\end{equation}

The output radar features $\mathcal{F}^{\text{Radar}}$ are resolution-aligned with the camera BEV features, ensuring seamless integration during multimodal fusion.

\subsection{Intensity-Aware Feature Fusion}

To demonstrate that intensity-aware mechanisms are not limited to LiDAR, we introduce an intensity-aware fusion strategy between camera and radar features. This approach enables modality-aware weighting during fusion, reducing the dominance of camera features and enhancing the contribution of radar in ambiguous or occluded regions.

Radar intensity is computed using both the Radar Cross Section (RCS) and Doppler velocity magnitude. For each radar point $i$, given its RCS value $\text{RCS}_i$ and Doppler velocity components $(v_{x_i}, v_{y_i})$, the intensity is defined as:
\begin{equation}
\label{equation6}
   \mathcal{I}_i^{\text{Radar}} = \sigma \left( \alpha \cdot \text{RCS}_i + \beta \cdot \sqrt{v_{x_i}^2 + v_{y_i}^2} \right)
\end{equation}
where $\alpha$ and $\beta$ are fixed scalar weights, and $\sigma$ denotes the sigmoid activation function.

For the camera features, a spatial intensity map $ \mathcal{I}^{\text{Cam}} \in \mathbb{R}^{N \times 1 \times H \times W}$ is generated using a convolutional layer:
\begin{equation}
\label{equation7}
    \mathcal{I}^{\text{Cam}} = \sigma(\text{Conv}(\mathcal{F}^{\text{Camera}}))
\end{equation}

We adopt a deformable attention formulation in BEV space, where radar BEV features \( \mathcal{F}^{\text{Radar}} \) serve as queries (\( \mathbf{Q} \)), and camera BEV features \( \mathcal{F}^{\text{Camera}} \) serve as keys (\( \mathbf{K} \)) and values (\( \mathbf{V} \)). The fusion is computed as:
\begin{align}
\label{equation8}
    \mathcal{F}^{\text{fused}} = \text{DeformAttn}_{\text{intensity}}(\mathbf{Q} = \mathcal{F}^{\text{Radar}},\ \mathbf{K} = \mathcal{F}^{\text{Camera}},\ \notag \\ \mathbf{V} = \mathcal{F}^{\text{Camera}},\ \mathcal{I}^{\text{Cam}},\ \mathcal{I}^{\text{Radar}})
\end{align}

This mechanism allows the network to learn spatially-varying cross-modal weights, reducing dominance from any single modality and improving the relevance of fused features.


\subsection{Adaptive Intensity-Guided Radar Feature Enhancement}

\begin{figure}[!h]
  \centering
    \includegraphics[width=1\linewidth]{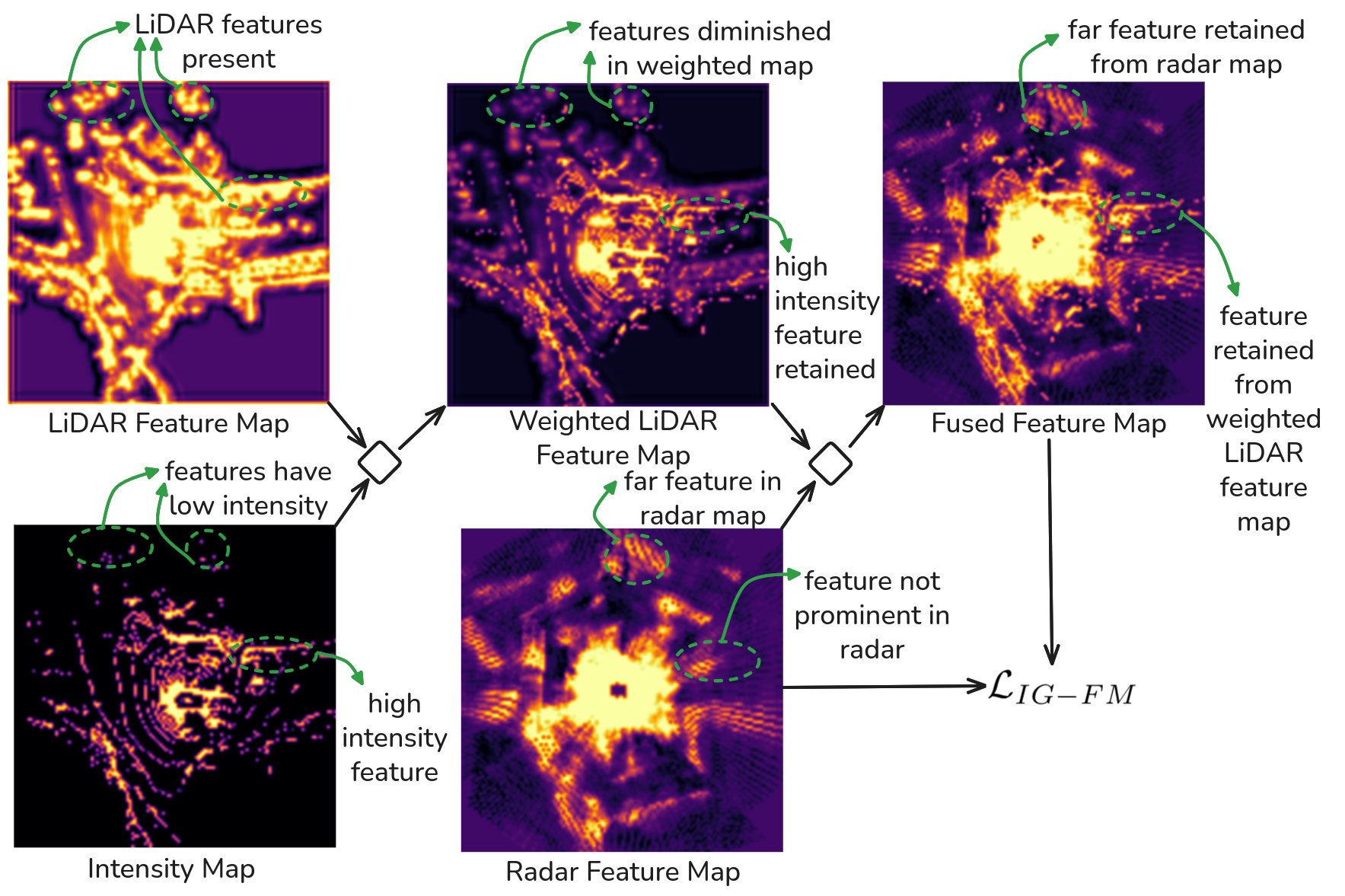}
   \caption{We illustrate a weighted LiDAR feature map, generated from intensity, is merged with the radar feature map to compute the intensity-guided feature map loss, preserving radar features and avoiding low-intensity LiDAR regions.}
   \label{fig:intensity_loss}
\end{figure}

While LiDAR intensity primarily reflects surface reflectivity, it also acts as a proxy for geometric confidence; high-intensity returns often correspond to structured and reflective objects such as vehicles or road boundaries. Rather than using intensity as a semantic indicator, we leverage it to prioritize supervision from high-confidence LiDAR regions during knowledge distillation. Crucially, instead of directly relying on raw intensities, we learn a transformation over the intensity map, enabling the network to adaptively reweight spatial contributions as shown in \cref{fig:intensity_loss}. 

\subsubsection{Voxelized LiDAR Intensity Extraction:} 
To derive intensity-aware guidance, we first voxelize LiDAR points 
\(\mathcal{P}_{\text{LiDAR}} = \{(x_i, y_i, z_i, I_i, t_i)\}_{i=1}^{N}\), 
where each point contains spatial coordinates \((x_i, y_i, z_i)\), intensity \(I_i\), and timestamp \(t_i\). 
Using a voxelization function \(\mathcal{V}\), we partition the points into discrete 3D voxels and compute the mean voxel intensity \(I_v\) as:
\begin{equation}
\label{equation9}
I_v = \frac{1}{N_v} \sum_{i=1}^{N_v} I_i,
\end{equation}

where \(N_v\) denotes the number of points within a voxel. This intensity is then projected onto a BEV grid \(\mathcal{I}_{\text{LiDAR}} \in \mathbb{R}^{H \times W}\), producing an intensity map aligned with the LiDAR BEV features:
\begin{equation}
\label{equation10}
\mathcal{I}^{\text{LiDAR}}(u, v) = \frac{\sum_{i} I_v \cdot \delta(u - u_i, v - v_i)}{\max(1, \sum_{i} \delta(u - u_i, v - v_i))},
\end{equation}

where \((u, v)\) are BEV grid indices from voxel projection, and \(\delta\) denotes voxel-to-grid mapping for spatial alignment.


\subsubsection{LiDAR-Weighted Radar Feature Fusion:} 
We utilize the intensity map \(\mathcal{I}^{\text{LiDAR}}\) as an adaptive weighting mechanism to guide the fusion of LiDAR and radar features. Given LiDAR feature map \(\mathcal{F}^{\text{LiDAR}}\) and radar feature map \(\mathcal{F}^{\text{Radar}}\), we compute intensity-based feature blending weights as:
\begin{equation}
\label{equation11}
w_{\text{LiDAR}} = \lambda \cdot \mathcal{I}^{\text{LiDAR}}, \quad w_{\text{Radar}} = 1 - w_{\text{LiDAR}},
\end{equation}

where \(\lambda\) is a learnable scaling factor. The fused BEV feature map is then constructed as:
\begin{equation}
\label{equation12}
\tilde{\mathcal{F}}^{\text{Radar}} = w_{\text{LiDAR}} \cdot \mathcal{F}^{\text{LiDAR}} + w_{\text{Radar}} \cdot \mathcal{F}^{\text{Radar}}.
\end{equation}

\subsubsection{Intensity-Aware Feature Map Loss for Radar Guidance:} 
To reinforce radar feature refinement, we introduce an alignment loss between radar features and the LiDAR feature map to encourage consistency between the two modalities:
\begin{equation}
\label{equation13}
\mathcal{L}_{\text{align}} = \left\| \mathcal{F}^{\text{Radar}} - \mathcal{F}^{\text{LiDAR}} \right\|^2.
\end{equation}

Simultaneously, we maintain a consistency loss between the radar and fused feature maps to prevent over-suppression of radar-specific information:
\begin{equation}
\label{equation14}
\mathcal{L}_{\text{consist}} = \left\| \mathcal{F}^{\text{Radar}} - \tilde{\mathcal{F}}^{\text{Radar}} \right\|^2.
\end{equation}

The final Intensity-Guided Feature Map Loss is formulated as:
\begin{equation}
\label{equation15}
\mathcal{L}_{\text{IG-FM}} = \alpha \mathcal{L}_{\text{align}} + (1 - \alpha) \mathcal{L}_{\text{consist}},
\end{equation}

where \(\alpha\) balances alignment and consistency constraints.

\subsection{LiDAR-Guided Feature Enhancement}
We directly supervise the fused camera–radar BEV representation using LiDAR as a privileged modality.
This process is challenged by semantic gaps between LiDAR and fused features, temporal misalignment across modalities, and instability from the high-dimensional, semantically enriched representation. To ensure stable and spatially-aware transfer, we apply LiDAR intensity as soft attention instead of binary or uniform weighting.

\subsubsection{Spatially-Weighted Feature Distillation}
We guide the fused BEV feature $\mathcal{F}^{\text{fused}} \in \mathbb{R}^{C \times H \times W}$ using LiDAR features $\mathcal{F}^{\text{LiDAR}}$ with spatial weighting from normalized LiDAR intensity $\mathcal{I}^{\text{LiDAR}} \in \mathbb{R}^{1 \times H \times W}$. The distillation loss is:

\begin{equation}
\label{equation16}
    \mathcal{L}_{\text{SWFD}} = \left\langle \mathcal{I}^{\text{LiDAR}}_{ij} \cdot \left\| \mathcal{F}_{ij}^{\text{LiDAR}} - \beta(\mathcal{F}_{ij}^{\text{fused}}) \right\|_2^2 \right\rangle_{i,j}
\end{equation}
Here, $\beta(\cdot)$ is a lightweight alignment module. This formulation ensures feature transfer is stronger in high-confidence regions while preserving full gradient flow—something not possible with handcrafted binary masks.

\subsubsection{Spatially-Weighted Response Distillation}
 To further align the predictions of the fused detector with the LiDAR teacher, we introduce a Spatially-Weighted Response Distillation loss.

\begin{align}
\label{equation17}
\mathcal{L}_{\text{SWRD}} &= \left\langle \mathcal{I}^{\text{LiDAR}}_{ij} \cdot \big( \mathcal{L}_{cls} ( \mathit{h}^{\text{LiDAR}}, \mathit{h}^{\text{fused}} ) \right. \notag \\
&\hspace*{\fill} \left. + \mathcal{L}_{bbox} ( \mathit{b}^{\text{LiDAR}}, \mathit{b}^{\text{fused}} ) \big) \right\rangle_{i,j}
\end{align}


Here, $\mathcal{L}_{cls}$ and $\mathcal{L}_{bbox}$ denote the standard classification and bounding-box regression losses (as in CenterPoint \cite{yin2021center}), with $\mathit{h}$ representing the predicted class heatmap and $\mathit{b}$ representing predicted bounding-box coordinates. This improves upon prior works by avoiding uniform foreground masks and instead applying confidence-weighted distillation across the full spatial domain.

\subsection{Label-Based Knowledge Distillation}

We extend LabelDistill \cite{kim2024labeldistill} to operate directly on fused camera-radar BEV features rather than Camera-only features. This avoids reliance on modality-specific artifacts and ensures robust supervision over the multi-modal representation.

In contrast to the binary object masks used in \cite{kim2024labeldistill}, we introduce a soft Gaussian mask centered on ground-truth boxes to enable smooth, graded supervision. The final loss is:

\begin{equation} 
\label{equation18}
\mathcal{L}_{\text{LD}} =  \frac{\sum \left\| \mathcal{F}^{\text{label}} - \mathcal{F}^{\text{fused}} \right\|^2 \cdot \mathcal{M}_{\text{label}}^{\text{soft}}}{\sum \mathcal{M}_{\text{label}}^{\text{soft}} + \epsilon}
\end{equation}

Here, $\mathcal{F}^{\text{label}}$ is the label-encoded BEV feature and $\mathcal{M}_{\text{label}}^{\text{soft}}$ softly weights the loss around valid object regions. This fused, soft-masked formulation improves generalization while preserving clean label supervision.

\subsection{Overall Loss Function and Training Strategy}

The total training objective combines detection, depth estimation, and distillation terms as:

\begin{align}
\label{equation19}
\mathcal{L}_{\text{total}} =\; & \lambda_{1} \mathcal{L}_{\text{det}} + \lambda_{2} \mathcal{L}_{\text{depth}} + \lambda_{3} \mathcal{L}_{\text{IG-FM}} \notag \\
& \hspace*{\fill} + \lambda_{4} \mathcal{L}_{\text{SWFD}} + \lambda_{5} \mathcal{L}_{\text{SWRD}} + \lambda_{6} \mathcal{L}_{\text{LD}}
\end{align}

All components are jointly optimized, with frozen LiDAR and label encoders used only during training. The modular design enables efficient fusion learning without inference-time overhead.

%% file: sec/4_experiments.tex
\section{Experiments}
\label{sec:experiments}

\subsection{Experimental Setup}
\subsection*{Dataset and Evaluation Metrics}

We evaluate on the nuScenes dataset \cite{caesar2020nuscenes}, which includes LiDAR, radar, and camera data across 1,000 scenes (850 for training/validation, 150 for testing). \newline
\textbf{3D object detection} is assessed using official nuScenes \cite{caesar2020nuscenes} metrics: mAP, NDS, and five TP metrics: mATE, mASE, mAOE, mAVE, and mAAE.

All results are reported on the nuScenes \cite{caesar2020nuscenes} validation and test sets for fair comparison with prior work.


\begin{table*}[b]
    \centering
    \setlength{\tabcolsep}{2pt} 
    \renewcommand{\arraystretch}{1}
    \begin{tabular}{l|c|c|c|c|c|c|c|c|c|c|c}
        \hline
        Method & Input & KD & Backbone & Image Size & NDS $\uparrow$ & mAP$\uparrow$ & mATE$\downarrow$ & mASE$\downarrow$ & mAOE$\downarrow$ & mAVE$\downarrow$ & mAAE$\downarrow$ \\
        \hline
        UVTR \cite{li2022unifying} & C & L2C & R101 & 900$\times$1600 & 45.0 & 37.2 & 0.735 & 0.269 & 0.397 & 0.761 & 0.193 \\
        BEVDistill \cite{chen2022bevdistill} & C & LC2C & R50 & 640$\times$1600 & 45.7 & 38.6 & 0.693 & 0.264 & 0.399 & 0.802 & 0.199 \\
        UniDistill \cite{zhou2023unidistill} & C & L2C & R50 & 256$\times$704 & 37.8 & 26.5 & - & - & - & - & - \\
        BEVSimDet \cite{kim2024labeldistill} & C & LC2C$\lozenge$2C & SwinT & 256$\times$704 & 45.3 & 40.4 & 0.526 & 0.275 & 0.607 & 0.805 & 0.273 \\
        LabelDistill \cite{kim2024labeldistill} & C & LL$\lozenge$2C & R50 & 256$\times$704 & 52.8 & 41.9 & 0.582 & 0.258 & 0.413 & 0.346 & 0.220 \\
        DistillBEV \cite{kim2024labeldistill} & C & L2C$\lozenge$2C & R50 & 256$\times$704 & 41.6 & 34.0 & 0.704 & 0.266 & 0.556 & 0.815 & 0.201 \\
        X3KD \cite{klingner2023x3kd} & C & LC2C & R50 & 256$\times$704 & 50.5 & 39.0 & 0.615 & 0.269 & 0.471 & 0.345 & 0.203 \\
        \midrule
        X3KD \cite{klingner2023x3kd} & C+R & L2CR & R50 & 256$\times$704 & 53.8 & 42.3 & - & - & - & - & - \\
        CRKD \cite{zhao2024crkd} & C+R & LC2CR & R50 & 256$\times$704 & 57.3 & 46.7 & 0.446 & 0.263 & 0.408 & 0.331 & 0.162 \\
        \rowcolor[gray]{0.9} IMKD (Ours) & C+R & LL$\lozenge$2M & R50 & 256 × 704 & \textbf{61.0} & \textbf{51.6} & \textbf{0.444} &\textbf{0.259} & \textbf{0.384} & \textbf{0.229} & \textbf{0.160} \\
        \hline
    \end{tabular}
    \caption{Comparison of Knowledge distillation (KD) methods for 3D object detection results on the nuScenes~\cite{caesar2020nuscenes} val set. 'L', 'L$\lozenge$' 'C', 'R' and 'M' denote LiDAR, label, camera, radar and merged (camera+radar) inputs, respectively.}
    \label{tab:comparison_kd_val}
\end{table*}

\subsection*{Implementation Details}

We use pretrained CenterPoint \cite{yin2021center} as the LiDAR teacher with $(0.1\text{m}, 0.1\text{m}, 0.2\text{m})$ voxel size and adopt the label encoder from LabelDistill \cite{kim2024labeldistill}. The camera branch is based on BEVDepth \cite{li2023bevdepth} with efficient depth layers. Radar inputs use multi-sweep projections with Doppler and RCS normalization, processed into a polar-to-BEV feature map. Temporal fusion follows BEVFormer \cite{li2203bevformer} by accumulating four BEV frames at 1s intervals, ensuring causal inference.

We use an ImageNet-pretrained ResNet50 \cite{he2016deep} backbone, with input size $256 \times 704$, trained using AdamW \cite{loshchilov2017decoupled}. Data augmentations are applied across all modalities. Detailed architectural and training configurations are provided in the supplementary material.

\subsection{Main Results and Comparison with State-of-the-Art}

\begin{table*}[t]
    \centering
    \setlength{\tabcolsep}{3pt} 
    \renewcommand{\arraystretch}{1}
    \begin{tabular}{l|c|c|c|c|c|c|c|c|c}
        \hline
        Method & Input & KD & NDS $\uparrow$ & mAP $\uparrow$ & mATE $\downarrow$ & mASE $\downarrow$ & mAOE $\downarrow$ & mAVE $\downarrow$ & mAAE $\downarrow$ \\
        \hline
        UVTR \cite{li2022unifying} & C & L2C & 52.2 & 45.2 & 0.612 & 0.256 & 0.385 & 0.664 & 0.125 \\
        BEVDistill \cite{chen2022bevdistill} & C & LC2C & 59.4 & 49.8 & 0.472 & 0.247 & 0.378 & 0.326 & 0.125 \\
        UniDistill \cite{zhou2023unidistill} & C & L2C & 39.3 & 29.6 & 0.637 & 0.257 & 0.492 & 1.084 & 0.167 \\
        LabelDistill \cite{kim2024labeldistill} & C & LL$\lozenge$2C & 61.0 & 52.6 & 0.443 & 0.252 & 0.339 & 0.370 & 0.136 \\
        X3KD \cite{klingner2023x3kd} & C & LC2C & 56.1 & 45.6 & 0.506 & 0.253 & 0.414 & 0.366 & 0.131 \\
        \midrule
        X3KD \cite{klingner2023x3kd} & C+R & L2CR & 55.3 & 44.1 & - & - & - & - & - \\
        CRKD \cite{zhao2024crkd} & C+R & LC2CR & 58.7 & 48.7 & 0.404 & 0.253 & 0.425 & 0.376 & 0.111 \\
        \rowcolor[gray]{0.9} IMKD (Ours) & C+R & LL$\lozenge$2M & \textbf{67.0} & \textbf{61.0} & \textbf{0.401} & \textbf{0.249} & \textbf{0.305} & \textbf{0.238} & \textbf{0.102} \\
        \hline
    \end{tabular}
    \caption{
        Comparison of Knowledge Distillation (KD) methods for 3D object detection on the nuScenes~\cite{caesar2020nuscenes} test set. 'L', 'L$\lozenge$', 'C', 'R', and 'M' denote LiDAR, label, camera, radar, and merged (camera+radar) inputs, respectively. 
        \textit{Note: Each method uses its best reported backbone and image size; comparisons should focus on distillation strategies.}
    }
    \label{tab:comparison_kd_test}
    \vspace{-0.3cm}
\end{table*}

We evaluate IMKD on the nuScenes \cite{caesar2020nuscenes} validation and test sets, comparing against a range of distillation-based 3D object detectors, including both camera-only and camera-radar student models. Results are summarized in Tables~\ref{tab:comparison_kd_val} and \ref{tab:comparison_kd_test}.

On the validation set, IMKD achieves 61.0 NDS and 51.6 mAP, outperforming all prior KD-based methods. Notably, it surpasses the strongest baseline, CRKD, by +6.5\% NDS and +10.1\% mAP, along with consistent reductions in translation, scale, and orientation errors. These gains highlight the effectiveness of IMKD’s fusion-level distillation and intensity-aware supervision.

On the nuScenes \cite{caesar2020nuscenes} test set, IMKD sets a new benchmark with 67.0 NDS and 61.0 mAP, significantly improving upon previous knowledge-distillation-based results. These improvements affirm that enhancing the quality and contextual relevance of supervision, rather than relying on modality-specific or heuristic KD, is key to unlocking robust camera-radar perception. Overall, these results demonstrate that IMKD offers a robust and generalizable approach to multi-modal knowledge distillation, advancing the field beyond conventional KD strategies and establishing a new standard for student models in 3D detection.

\begin{figure*}[t]
  \centering
    \includegraphics[width=0.9\linewidth]{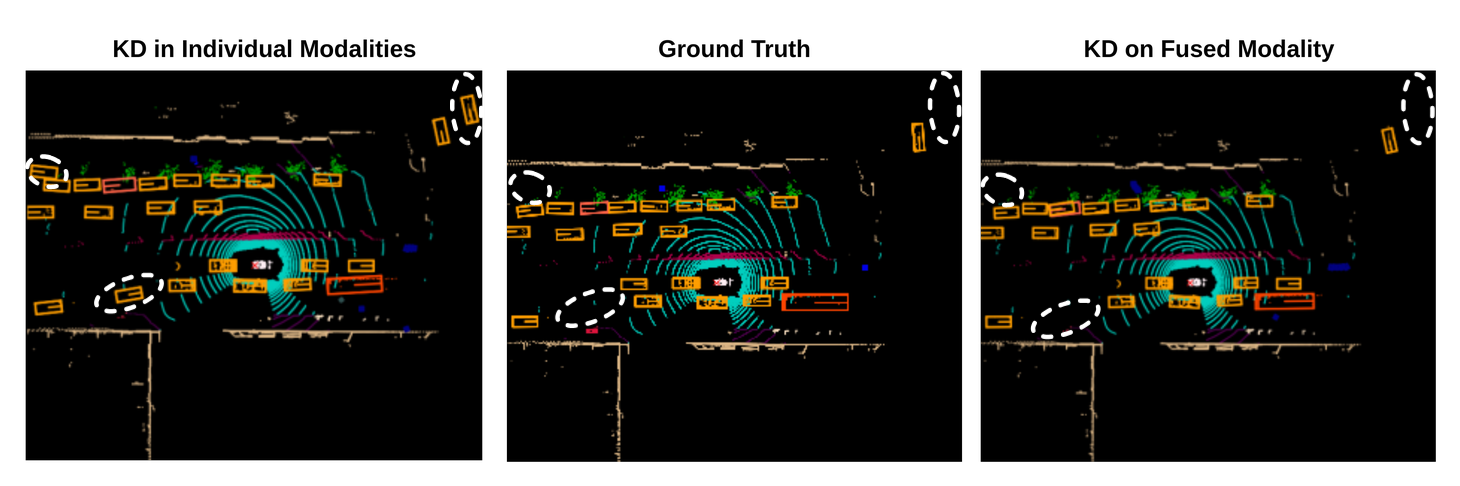}
    \vspace{-0.4cm}
   \caption{Comparison of distillation targets: individual modality KD yields extra false detections (white circles) and poor orientation, while merged feature KD aligns better with ground truth.}
   \label{fig:kd_diff}
   \vspace{-0.3cm}
\end{figure*}

Qualitative results in Fig.~\ref{fig:kd_diff} further illustrate IMKD’s impact: knowledge distillation on fused modalities yields more accurate detections and better box orientation compared to individual-modality KD, especially in ambiguous or occluded scenes.

In the following section \ref{sec:ablation}, we present ablation studies to analyze the contributions of each IMKD component. Unless otherwise specified, all experiments are conducted on the nuScenes \cite{caesar2020nuscenes} validation set using a ResNet-50 \cite{he2016deep} image backbone. We report standard metrics including mAP, NDS, and detailed error breakdowns to evaluate performance comprehensively.

\subsection{Ablation Study}
\label{sec:ablation}
We conduct a step-wise ablation to isolate the impact of each component in IMKD. Starting from a camera-only baseline, we progressively introduce radar grid learning, intensity-aware fusion, and our distillation modules.

\subsubsection{Effectiveness of Learnable Radar Grid}
We assess the effect of radar input design on detection performance in \cref{tab:radar_grid_cmparison}. Introducing radar via a fixed handcrafted grid boosts performance over the camera-only setup, confirming the benefit of multi-modal fusion. Switching to a learnable radar grid further improves both mAP and NDS, validating its role in producing task-adaptive radar features.
\begin{table}[H]
    \centering
    \renewcommand{\arraystretch}{0.9} 
    \setlength{\tabcolsep}{1pt} 
    \begin{tabular}{l| c| c}
        \toprule
        Method & mAP $\uparrow$ & NDS $\uparrow$ \\
        \midrule
        Camera Only & 34.8 & 44.6 \\
        Camera+Radar (Handcrafted) & 41.2 & 52.5 \\
        Camera+Radar (Learnable) - Baseline & 43.4 & 53.5 \\
        \bottomrule
    \end{tabular}
    \caption{Ablation on radar representation. Learnable radar grid improves over both camera-only and handcrafted radar projection.}
    \label{tab:radar_grid_cmparison}
    \vspace{-0.2cm}
\end{table}

\subsubsection{Impact of Intensity-Aware C+R Fusion}
We incorporate both radar and camera intensity to guide the fusion process. This leads to notable gains over the learnable grid baseline as shown in \cref{tab:single_vs_merged}, validating the benefit of confidence-aware fusion. Importantly, this shows that intensity-guided processing is beneficial beyond LiDAR and can be generalized to radar-camera fusion.
\begin{table}[H]
    \centering
    \renewcommand{\arraystretch}{0.9} 
    \setlength{\tabcolsep}{2pt} 
    \begin{tabular}{l| c| c}
        \toprule
        Method & mAP $\uparrow$ & NDS $\uparrow$ \\
        \midrule
        Baseline &  43.4 & 53.5 \\
        + Intensity-Aware Fusion & 46.5 & 55.3 \\
        \bottomrule
    \end{tabular}
    \caption{Comparison of non-intensity-based vs. intensity-aware fusion for merged camera-radar features.}
    \label{tab:single_vs_merged}
    \vspace{-0.2cm}
\end{table}

\subsubsection{Effectiveness of Proposed KD Modules}
We incrementally evaluate the effectiveness of our proposed distillation objectives on top of the intensity-aware fusion baseline in \cref{tab:ablation_losses}. Each module, i.e. LiDAR feature distillation, label distillation, intensity-guided feature map supervision, and response distillation yields consistent performance gains, validating their individual contribution. When combined, they offer additive improvements, with the full IMKD model achieving a notable +11\% mAP and +9.3\% NDS over the baseline. This demonstrates the effectiveness of our modular, fusion-aligned KD framework in enhancing multi-modal perception.
\begin{table}[H]
    \centering
    \renewcommand{\arraystretch}{1}
    \setlength{\tabcolsep}{2pt}
    \begin{tabular}{l|c|c}
    \toprule
    \textbf{Configuration} & \textbf{mAP} $\uparrow$ & \textbf{NDS} $\uparrow$ \\
    \midrule
    Intensity-Aware Fusion & 46.5 & 55.3 \\
    + LiDAR Feature Distill ($\mathcal{L}_{\text{SWFD}}$) & 49.56 & 58.30 \\
    + Label Distill ($\mathcal{L}_{\text{LD}}$) & 49.41 & 58.25 \\
    + Intensity-Guided Feature Map ($\mathcal{L}_{\text{IG-FM}}$) & 49.64 & 58.40 \\
    + Response Distill ($\mathcal{L}_{\text{SWRD}}$) & 47.68 & 57.31 \\
    \midrule
    + Response + LiDAR Distill. & 49.72 & 58.51 \\
    + Resp. + LiDAR + Label Distill. & 50.22 & 59.14 \\
    IMKD (Full Model) & \textbf{51.65} & \textbf{61.05} \\
    \bottomrule
    \end{tabular}
    \caption{Ablation on proposed distillation objectives over the intensity-guided fusion baseline.}
    \label{tab:ablation_losses}
    \vspace{-0.4cm}
\end{table}

\subsubsection{Cross-Modal vs Uni-Modal Distillation}
We compare distillation into individual modalities versus fused features. While unimodal KD achieves solid results, fusion-level KD yields +4.0\% mAP and +4.6\% NDS as shown in \cref{tab:ablation_fusion_kd}. These gains require mitigating gradient conflicts and applying pseudo-label masking and feature normalization, underscoring the complexity of fusion-level supervision. This validates our hypothesis that modality interaction within the distillation target space plays a crucial role in enhancing downstream performance.
\begin{table}[H]
    \vspace{-0.3cm}
    \centering
    \renewcommand{\arraystretch}{1}
    \setlength{\tabcolsep}{4pt}
    \begin{tabular}{l|l|c|c}
    \toprule
    LiDAR KD Target & Label KD Target & mAP $\uparrow$ & NDS $\uparrow$ \\
    \midrule
    Camera \& Radar & Camera \& Radar & 49.8 & 58.3 \\
    Fused & Camera \& Radar & 50.5 & 59.3 \\
    Camera \& Radar & Fused & 50.2 & 59.0 \\
   Fused & Fused & \textbf{51.6} & \textbf{61.0} \\
    \bottomrule
    \end{tabular}
    \caption{Comparison of distillation targets using individual modalities vs. fused camera-radar features.}
    \label{tab:ablation_fusion_kd}
    \vspace{-0.3cm}
\end{table}

\subsection{Robustness to Visibility and Temporal Degradation}
To our knowledge, IMKD is the first distillation-based fusion framework to evaluate performance under diverse environmental conditions (e.g., rain, night) and degraded temporal input (frame drop). Prior KD-based methods such as CRKD \cite{zhao2024crkd}, X3KD \cite{klingner2023x3kd}, and LabelDistill \cite{kim2024labeldistill} focus solely on standard benchmarks, leaving real-world robustness unexplored. Our analysis in \cref{tab:lighting_weather} reveals that IMKD exhibits greater stability in adverse conditions, suggesting that confidence-aware distillation leads to more reliable multi-modal fusion. Although IMKD uses LiDAR during training, it remains effective even when supervision is noisy or partially missing, as the learned guidance is intensity-adaptive and spatially grounded, rather than hard-coded. To contextualize these results, we also report robustness for non-KD fusion methods (CRN \cite{kim2023crn}, RCBEV \cite{zhou2023bridging}). While not directly comparable, they offer a useful reference point for deployment. IMKD consistently maintains superior performance across weather and frame drop scenarios \cref{tab:view_drop_comparison}, demonstrating that confidence-guided distillation improves both accuracy and resilience in safety-critical conditions.

\begin{table}[h]
    \centering
    \renewcommand{\arraystretch}{1} 
    \setlength{\tabcolsep}{3pt} 
    \begin{tabular}{l|c|c|c|c|c}
        \toprule
        Input & Modality & Sunny & Rainy & Day & Night \\
        \midrule
        RCBEV \cite{zhou2023bridging} & C+R & 36.1 & 38.5 & 37.1 & 15.5 \\
        BEVDepth \cite{li2023bevdepth} & C & 39.0 & 39.0 & 39.3 & 16.8 \\
        CRN \cite{kim2023crn} & C+R & 54.8 & 57.0 & 55.1 & 30.4 \\
        \rowcolor[gray]{0.9} IMKD (Ours) & C+R & \textbf{57.9} & \textbf{58.5} & \textbf{58.3} &\textbf{34.7} \\
        \bottomrule
    \end{tabular}
    \vspace{-0.5em}
    \caption{mAP under varying weather and lighting on nuScenes \cite{caesar2020nuscenes} val set, where IMKD outperforms other methods.}
    \label{tab:lighting_weather}
    \vspace{-0.4cm}
\end{table}

\begin{table}[H]
    \centering
    \renewcommand{\arraystretch}{1}
    \setlength{\tabcolsep}{1.5pt}
    \begin{tabular}{l|c|c|c|c|c|c}
        \toprule
        Method & Input & Modality & Drop 0 & Drop 1 & Drop 3 & Drop 6 \\
        \midrule
        \multirow{2}{*}{CRN \cite{kim2023crn}} & \multirow{2}{*}{C+R} & \multirow{2}{*}{\begin{tabular}{@{}c@{}} C \\ R \end{tabular}} & \multirow{2}{*}{47.18} & \multirow{2}{*}{\begin{tabular}{@{}c@{}} 40.19 \\ 45.39 \end{tabular}} & \multirow{2}{*}{\begin{tabular}{@{}c@{}} 22.94 \\ 41.34 \end{tabular}} & \multirow{2}{*}{\begin{tabular}{@{}c@{}} - \\ 34.52 \end{tabular}} \\
        & & & & & \\
        \midrule
        \multirow{2}{*}{IMKD} & \multirow{2}{*}{C+R} & \multirow{2}{*}{\begin{tabular}{@{}c@{}} C \\ R \end{tabular}} & \multirow{2}{*}{\textbf{51.6}} & \multirow{2}{*}{\begin{tabular}{@{}c@{}} \textbf{43.28} \\ \textbf{49.47} \end{tabular}} & \multirow{2}{*}{\begin{tabular}{@{}c@{}} \textbf{23.53} \\ \textbf{44.40} \end{tabular}} & \multirow{2}{*}{\begin{tabular}{@{}c@{}} - \\ \textbf{36.35} \end{tabular}} \\
        & & & & & \\
        \bottomrule
    \end{tabular}
    \vspace{-0.5em}
    \caption{ mAP under increasing frame drops on nuScenes \cite{caesar2020nuscenes} val set, where IMKD remains more stable across sensor degradations.}
    \label{tab:view_drop_comparison}
\end{table}

%% file: sec/5_conclusion.tex
\vspace{-2em}
\section{Conclusion}\label{sec:conclusion}
We proposed IMKD, an Intensity-guided Multi-Level Knowledge Distillation framework for radar-camera 3D object detection. IMKD identifies a core limitation in existing multi-modal distillation methods: modality-specific supervision often leads to incoherent representations and suboptimal fusion. To address this, we introduce a merged feature distillation strategy and an intensity-aware refinement module that prioritizes high-confidence regions during training. Although IMKD is trained using LiDAR as a privileged modality, its design is agnostic to specific sensor pairs and may be extended to other domains where confidence-guided supervision is available. Through extensive ablations and comparisons on nuScenes \cite{caesar2020nuscenes}, we demonstrated that IMKD delivers consistent improvements over state-of-the-art distillation baselines, validating both the design and effectiveness of our proposed framework. As a future direction, we plan to extend IMKD to temporal multi-frame fusion, enabling dynamic scene understanding and improved consistency in long-term predictions.

%% file: sec/acknowledgement.tex
\section{Acknowledgements}
This work was partially funded by the European Union under Grant Agreement No. 101076360 (BERTHA) and by the German Federal Ministry of Research, Technology and Space (BMFTR) under Grant Agreement No. 01IW24009 (COPPER).

%% file: sec/6_supplementary.tex
\clearpage
\setcounter{page}{1}
\twocolumn[
\begin{center}
    \LARGE \textbf{IMKD: Intensity-Aware Multi-Level Knowledge Distillation for Camera-Radar Fusion} \\
    \LARGE {Supplementary Material}
\end{center}
\vspace{1em} 
]
\label{sec:supplementary}
\section{Overview}
This supplementary material provides additional details on our proposed approach, including architectural design choices, implementation specifics, and extended experimental results. In \cref{supp:Arch}, we describe the network architecture and key design decisions. \cref{supp:Implementation} covers implementation details, including data preprocessing, hyperparameters, training configuration, and the impact of feature partitioning on fusion. In \cref{supp:Exp_Results}, we present additional experimental results, such as per-class performance analysis. Finally, \cref{supp:Qualitative} provides qualitative results to further illustrate the effectiveness of our method.
\section{Architectural Details}\label{supp:Arch}
\subsection{Motivation for Intensity-Guided Distillation} \label{supp:intensity_motivation}
LiDAR intensity encodes the strength of signal returns, which is closely tied to geometric reliability and boundary consistency \cite{major2019vehicle, yin2021center, lang2019pointpillars}. In IMKD, intensity is not transferred as a raw feature; instead, it is used to guide knowledge distillation and fusion. Specifically, LiDAR supervision is intensity-weighted when transferring features to the camera–radar fused representation, while camera and radar intensities are also used to modulate their deformable fusion. This ensures that distillation emphasizes reliable LiDAR regions, aligns multi-sensor features, and sharpens fused predictions.  

The motivation is threefold: intensity provides a reliability prior that (i) emphasizes consistent LiDAR features during transfer, preventing noisy regions from dominating; (ii) improves alignment of camera–radar fusion by highlighting structurally meaningful areas for cross-modal attention; and (iii) refines prediction confidence by guiding the fused BEV representation toward sharper object boundaries and more stable detections.  

This design avoids directly forcing radar to mimic LiDAR, preserving radar’s modality-specific robustness (e.g., under adverse weather), while still leveraging LiDAR’s depth-rich supervision. Similar strategies have proven effective beyond autonomous driving. In Medical imaging, MRI and CT often leverage intensity-weighted priors to guide segmentation, where voxel intensity correlates with tissue density and boundary sharpness \cite{litjens2017survey,ronneberger2015u}.  
In Remote sensing, satellite imagery, and radar backscatter intensity is used to enhance feature fusion for land-cover classification and flood detection, where high-return regions correspond to structurally reliable terrain \cite{huang2024deep,shafique2022deep}.  

These parallels show that intensity is a widely validated proxy for reliability and structure across domains. By incorporating it into the distillation process, IMKD enhances the transfer of geometric and structural knowledge without erasing modality-specific strengths.


\subsection{Architectural Design Considerations}\label{supp:Arch_Design}

Our architecture is designed with modularity and supervision efficiency in mind. While the main paper details the overall pipeline, here we highlight key considerations behind specific design choices that enhance robustness and enable clean integration of privileged signals.

\textbf{Intensity-Aware Cross-Modality Fusion:}
Our architecture fuses camera and radar features using a deformable attention mechanism guided by both camera confidence and radar intensity maps. This dual-intensity guidance enables the network to adaptively align features across modalities, prioritizing reliable regions and suppressing noise. Unlike modality-agnostic or uniform fusion schemes, our design selectively emphasizes trustworthy cues from each sensor, leading to more robust representations under challenging conditions such as rain, night, or partial sensor failure.

In Eq.~\ref{equation8}, the camera and radar intensity maps $\mathcal{I}^{\text{Cam}}, \mathcal{I}^{\text{Radar}}$ serve as modulation signals within the deformable attention module.
Specifically, intensity values are concatenated with key–value embeddings and passed through a learned gating function $g(\cdot)$, which rescales both the sampling offsets and the attention weights:
\begin{align}
w_{ij} = \text{softmax}\big( (\mathbf{q}_i \cdot \mathbf{k}_j) \cdot g(\mathcal{I}_j) \big),
\end{align}
where $g(\cdot)$ is a lightweight MLP with sigmoid activation. This mechanism ensures that attention is biased toward high-intensity regions (i.e., geometrically reliable points), enabling intensity-aware feature selection and fusion.

\textbf{Intensity-Guided Radar Representation:}
Radar intensity is used to modulate the radar branch features before fusion. Although simple, this plays a vital role in enhancing geometric priors, especially under sensor degradation (e.g., frame drops or poor lighting). This design avoids the need for radar-specific heuristics or handcrafted filters.

\textbf{Late Injection of Supervision:}
All remaining distillation losses are injected post-fusion, reducing the risk of modality dominance and preserving the integrity of radar features during training. This ensures that supervision acts as a guidance mechanism, not a constraint.

\textbf{Drop-In Extensibility:}
The design is easily extendable to other sensor pairs, e.g., camera+thermal or camera+event. Our use of post-fusion supervision and intensity-aware enhancement ensures that new modalities can be added without major architectural changes.

These choices, while not architectural novelties in isolation, collectively enable IMKD to scale well under different conditions and sensor setups with minimal adjustments.

\subsection{Inference Pipeline}
During inference, our model operates efficiently using only camera and radar inputs, ensuring a lightweight and deployable architecture. Several components used during training are discarded, streamlining computation without compromising detection performance.

Components Removed at Inference:

\textbf{LiDAR Feature Maps:} Since LiDAR supervision is only utilized during training to inject spatial priors, these feature maps are not required at test time.

\textbf{Label Encoder:} The label encoder, responsible for transforming ground truth 3D bounding boxes into a BEV representation, is used solely for training supervision and is omitted during inference.

\textbf{Efficient Operation with Camera and Radar Inputs:}
At inference, multi-view camera and radar features are first projected into a BEV space. These BEV features are then fused using an intensity-aware deformable fusion module, which leverages both camera confidence scores and radar intensity maps to guide spatial alignment. This design ensures robustness under adverse conditions by emphasizing high-confidence regions from each modality. Although LiDAR and label supervision are used during training, they are not required at test time. As a result, our method achieves accurate and efficient 3D detection using only camera and radar inputs, making it practical for real-world deployment.

\subsection{Inference Time}
We evaluate the inference speed of our IMKD framework on an RTX 3090 GPU using a single batch with FP16 precision. With a ResNet-50 \cite{he2016deep} backbone, our method achieves real-time performance at 25 FPS, making it competitive with existing camera-radar fusion approaches. Our knowledge distillation framework is employed solely during training and introduces no additional latency during inference.

Among knowledge distillation-based 3D detection methods, only BEVSimDet \cite{zhao2024simdistill} and UVTR-C \cite{li2022unifying} report inference speeds—11.1 FPS and 3.1 FPS, respectively—while BEVDet-Tiny \cite{huang2021bevdet} (a camera-only baseline) runs at 15.6 FPS. Other methods, such as UniDistill \cite{zhou2023unidistill}, LabelDistill \cite{kim2024labeldistill}, DistillBEV \cite{chen2022bevdistill}, X3KD \cite{klingner2023x3kd} and CRKD \cite{zhao2024crkd}, do not disclose inference performance. In contrast, our IMKD model delivers 25 FPS while outperforming these methods in detection accuracy, highlighting its strong balance between efficiency and robustness for real-world deployment.

\begin{table}[h]
    \centering
    \renewcommand{\arraystretch}{1}
    \setlength{\tabcolsep}{8pt}
    \begin{tabular}{l|c|c}
    \toprule
    Method & Type & FPS \\
    \midrule
    BEVDet-Tiny \cite{huang2021bevdet}  & Camera-Only & 15.6 \\
    UVTR-C \cite{li2022unifying} & KD-Based & 3.1 \\
    BEVSimDet \cite{zhao2024simdistill} & KD-Based & 11.1 \\
    \rowcolor[gray]{0.9}
    \textbf{IMKD (Ours)} & KD-Based & \textbf{25.0} \\
    \bottomrule
    \end{tabular}
    \caption{Comparison of inference speeds (FPS) across KD-based and camera-only baselines. Our method achieves real-time performance while maintaining strong accuracy.}
    \label{tab:fps_comparison}
\end{table}

\subsection{Loss Function Weight Tuning} \label{supp:loss_tuning}

The weights for individual loss terms in \cref{equation19} are empirically tuned to ensure balanced contributions during training. Specifically, the detection and depth losses ($\lambda_1, \lambda_2$) are set to 0.3, while the LiDAR- and label-based distillation losses ($\lambda_4, \lambda_5, \lambda_6$ in \cref{equation16}, \cref{equation17}, and \cref{equation18}) are also weighted at 0.3 to provide auxiliary supervision without overwhelming the primary objectives. The radar distillation loss ($\lambda_3$ in \cref{equation15}) is governed by a learnable scalar, initialized at 100, which allows the network to adaptively adjust its relative contribution during training and reduces manual sensitivity. Within \cref{equation15}, the alignment–consistency trade-off is controlled by $\alpha = 0.5$, which provides a balanced emphasis across geometric consistency and feature alignment.

\begin{table}[H]
    \centering
    \renewcommand{\arraystretch}{1}
    \setlength{\tabcolsep}{2pt}
    \begin{tabular}{l|c|p{1.2cm}}
    \toprule
    \textbf{Loss Term} & \textbf{Symbol} & \textbf{Weight} \\
    \midrule
    Detection loss ($\mathcal{L}_{\text{det}}$) & $\lambda_1$ & 0.3 \\
    Depth loss ($\mathcal{L}_{\text{depth}}$) & $\lambda_2$ & 0.3 \\
    Intensity-Guided Feature Map ($\mathcal{L}_{\text{IG-FM}}$) & $\lambda_3$ & Learn., init.~100 \\
    LiDAR Feature Distill ($\mathcal{L}_{\text{SWFD}}$) & $\lambda_4$ & 0.3 \\
    Response Distill ($\mathcal{L}_{\text{SWRD}}$) & $\lambda_5$ & 0.3 \\
    Label Distill ($\mathcal{L}_{\text{LD}}$) & $\lambda_6$ & 0.3 \\
    Alignment-consistency trade-off & $\alpha$ & 0.5 \\
    \bottomrule
    \end{tabular}
    \caption{Loss functions and corresponding weights used in IMKD.}
    \label{tab:loss_weights}
\end{table}

\begin{figure*}[t]
  \centering
    \includegraphics[width=1\linewidth]{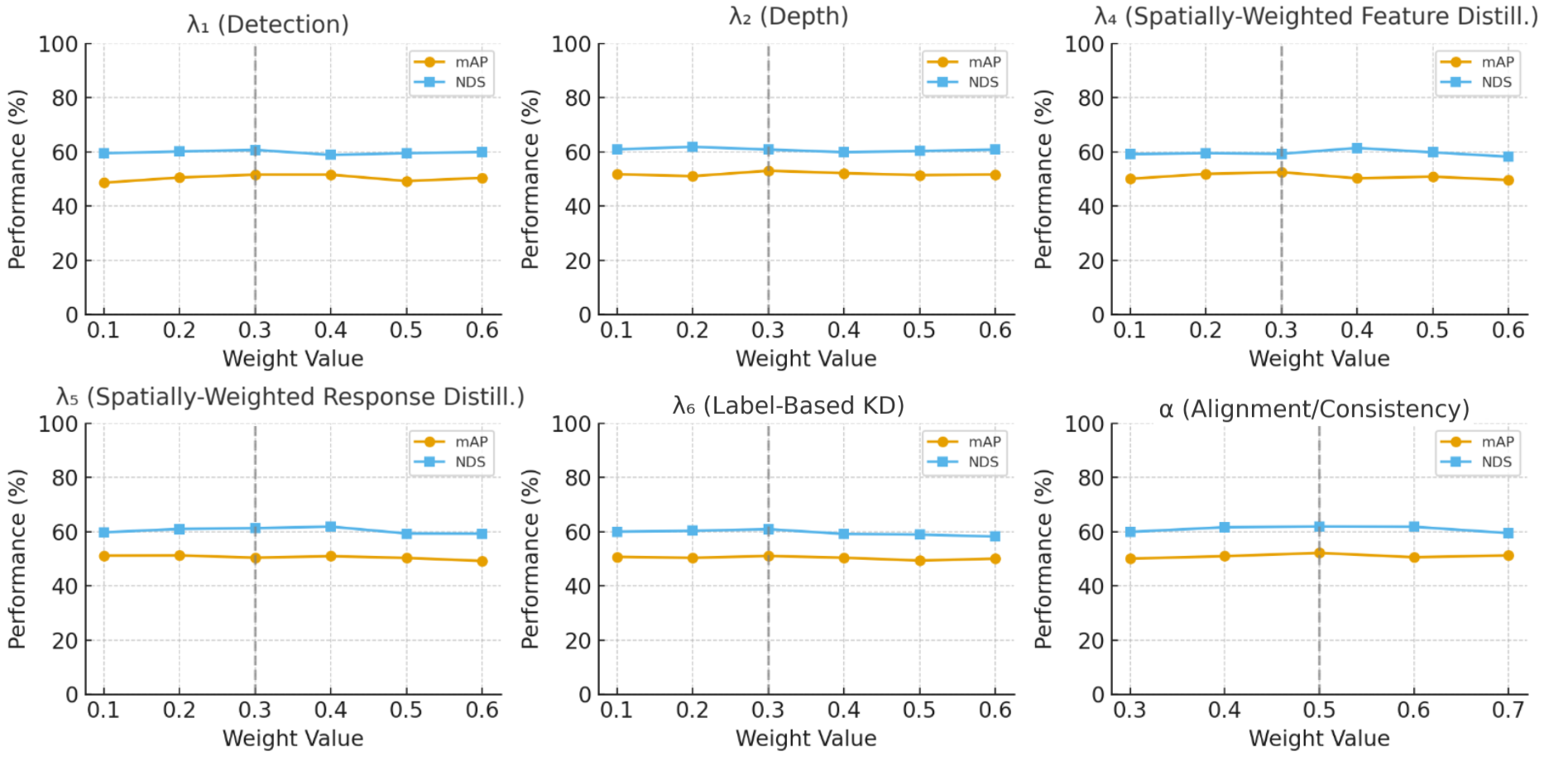}
   \caption{Sensitivity of mAP and NDS to individual loss weights $\lambda$. Each subplot reports an illustrative sweep over $\lambda \in (0.1,0.2,0.3,0.4,0.5,0.6)$; dashed vertical lines mark the chosen operating points ($\lambda=0.3$ for most terms, $\alpha=0.5$ for alignment). The curves indicate that performance is stable near the chosen weights and degrades when weights deviate substantially.}
   \label{fig:loss_graph}
\end{figure*}

These settings were chosen after preliminary sweeps to equalize the order of magnitude of gradients from each term, preventing instability from any single loss. We observed that training remained stable across all experiments without requiring further re-tuning, indicating that the framework is not overly sensitive to precise hyperparameter choices. The final values used in all experiments are summarized in \cref{tab:loss_weights} for reproducibility. This stability is also illustrated in \cref{fig:loss_graph}, where we plotted the values of all loss weights. The curves show that performance remains largely stable near the chosen weights, while substantial deviations can lead to degradation, confirming that the selected operating points strike a robust balance across losses.


\section{Implementation Details} \label{supp:Implementation}

\subsection{Pre-Processing}
\textbf{Pre-processing} \newline
Our method utilizes multi-modal data comprising images, radar, and LiDAR point clouds. The following pre-processing steps are applied to each modality:
\newline
\textbf{Image Pre-processing:} Images undergo a random resize within a scaling range of 
$[0.386,0.55]$ before being cropped to a fixed resolution of $256 \times 704$. Data augmentation includes random horizontal flipping and a constrained vertical crop with no bottom percentage limit. Rotation augmentation is disabled. Six camera views are used.
\newline
\textbf{Radar Pre-processing:} Radar points are projected into the BEV space, with an intensity-aware transformation applied to align them with the camera features. The radar representation is downsampled using a voxelization process with a fixed BEV grid resolution.
\newline
\textbf{LiDAR Pre-processing:} LiDAR points are voxelized with a voxel size of $[0.1,0.1,0.2]$, ensuring consistent spatial resolution. The voxel encoder uses a sparse convolutional network to generate a compact feature representation while maintaining high spatial fidelity.
\newline
\textbf{BEV Augmentation:} BEV-space transformations include a random rotation within $[-22.5\text{\textdegree},22.5\text{\textdegree}]$, a scaling perturbation in the range $[0.9,1.1]$, and a probabilistic flipping along both axes with a 50\% chance.
\subsection{Hyperparameters Settings}
\textbf{Backbone (Image Branch)}: A ResNet-50 \cite{he2016deep} extracts multi-scale image features, processed via an FPN-style \cite{lin2017feature} neck with an upsampling strategy of $\{0.25, 0.5, 1, 2\}$.  
\newline
\textbf{Backbone (Radar \& LiDAR Fusion)}: Point cloud features are voxelized and encoded using a SECOND-based \cite{yan2018second} architecture, followed by a stacked CNN backbone. The features are refined via a SECONDFPN-style neck with output strides of $\{0.5, 1, 2\}$.  
\newline
\textbf{Detection Head}: The detection follows a CenterPoint-style \cite{yin2021center} approach, leveraging a hierarchical BEV backbone and an FPN-style \cite{lin2017feature} neck. Bounding boxes are regressed using a CenterPoint-based \cite{yin2021center} box coder with a post-center range of $[-61.2, 61.2]$.

\subsection{Training Configuration}
\textbf{Loss Functions}: Apart from the losses mentioned in the paper, the classification loss is based on Gaussian Focal Loss \cite{wang2021gaussian}, while regression losses include L1 Loss \cite{hastie2009elements} for bounding box estimation and a smooth transition function for orientation prediction. Additional loss terms are incorporated to enhance knowledge-distillation and overall detection performance.
 \newline
\textbf{Voxelization}: The LiDAR point cloud is voxelized within a spatial range of $[-51.2, 51.2]$ meters in the XY plane and a vertical range from $-5$ to $3$ meters.  
\newline
\textbf{Training Grid Settings}: The BEV grid is constructed with a spatial resolution of $[512, 512]$ and an output downsampling factor of $4$. For LiDAR, the grid is defined over $[1024, 1024, 40]$ points, maintaining high spatial fidelity.

\begin{table}[h]
    \centering
    \renewcommand{\arraystretch}{1.2} 
    \begin{tabular}{l|c}
        \toprule
        Config & ResNet-50/101 \\
        \midrule
        Optimizer & AdamW \\
        Base Learning Rate & $4e-4$ \\
        Backbone Learning Rate & $2e-4 / 1e-4$ \\
        Weight Decay & $1e-2$ \\
        Batch Size & 16 / 8 \\
        Training Epochs & 30 \\
        LR Schedule & Cosine \\
        Gradient Clip & 5 \\
        \bottomrule
    \end{tabular}
    \caption{Training configurations for ResNet-50/101.}
    \label{tab:LR_configs}
\end{table}



\section{Additional Experimental Results} \label{supp:Exp_Results}

\subsection{Comparison with LiDAR Teacher Model} \label{supp:l_c_comparison}
To evaluate the effectiveness of IMKD, we compare it against its LiDAR-based teacher, specifically CenterPoint \cite{yin2021center} pretrained on the nuScenes \cite{caesar2020nuscenes} dataset. The student model consists of a BEVDepth \cite{li2023bevdepth} camera module and a radar encoder.

\cref{tab:teacher_comparison} summarizes the results. While the LiDAR teacher achieves strong performance, it is not the best-performing LiDAR model on the nuScenes \cite{caesar2020nuscenes} dataset. We report IMKD results with and without distillation. Although direct comparison across modalities is inherently challenging, distillation significantly improves the student, with NDS increasing by 1.8 and mAP by 2.6 compared to the teacher. This improvement arises because our multi-level distillation transfers depth cues, geometric structure, and point-density patterns from LiDAR into the fused camera–radar representation, thereby compensating for the modalities’ inherent weaknesses. In addition, the prediction-level distillation between LiDAR outputs and the student predictions refines decision boundaries and reduces ambiguity in challenging cases. Together, these mechanisms allow the student to not only close the gap with the LiDAR teacher but in some settings surpass it by leveraging complementary cross-modal information absent in LiDAR alone.

\begin{table}[h]
    \centering
    \renewcommand{\arraystretch}{1.2}
    \setlength{\tabcolsep}{8pt}
    \begin{tabular}{l|c|c}
    \toprule
    Method & mAP & NDS \\
    \midrule
    LiDAR Teacher \cite{yin2021center}  & 58.40 & 65.20 \\
    IMKD w/o LiDAR Distil. & 56.90 & 62.5 \\
    \rowcolor[gray]{0.9}IMKD Full & \textbf{61.0} & \textbf{67.0} \\
    \bottomrule
    \end{tabular}
    \caption{Performance comparison between our IMKD model and its LiDAR teacher on the nuScenes \cite{caesar2020nuscenes} test set.}
    \label{tab:teacher_comparison}
\end{table}

\subsection{Comparison with Camera-Radar Methods without Knowledge Distillation} \label{supp:c-r_comparison}
To further contextualize the performance of our IMKD framework, we compare it against recent camera-radar fusion methods that do not use knowledge distillation. As shown in Table~\ref{tab:comparison_val}, we benchmark IMKD against several strong baselines including CRN~\cite{kim2023crn}, RCBEVDet~\cite{lin2024rcbevdet}, and CRT-Fusion~\cite{kim2024crt}, all evaluated on the nuScenes \cite{caesar2020nuscenes} validation set.

\begin{table*}[t]
    \centering
    \setlength{\tabcolsep}{3.5pt} 
    \renewcommand{\arraystretch}{1.1}
    \begin{tabular}{l|c|c|c|c|c|c|c|c|c|c}
        \toprule
        Method & Input & Backbone & Image Size & NDS↑ & mAP↑ & mATE↓ & mASE↓ & mAOE↓ & mAVE↓ & mAAE↓ \\
        \midrule
        $\text{BEVDet}$ \cite{huang2021bevdet} & C & ResNet-50 & 256 × 704 & 39.2 & 31.2 & 0.691 & 0.272 & 0.523 & 0.909 & 0.247 \\
        $\text{BEVDepth}$ \cite{li2023bevdepth} & C & ResNet-50 & 256 × 704 & 47.5 & 35.1 & 0.639 & 0.267 & 0.479 & 0.428 & 0.198 \\
        RC-BEVFusion \cite{stacker2023rc} & C+R & ResNet-50 & 256 × 704 & 52.5 & 43.4 & 0.511 & 0.270 & 0.527 & 0.421 & 0.182 \\
        $\text{SOLOFusion}$ \cite{park2022time} & C & ResNet-50 & 256 × 704 & 53.4 & 42.7 & 0.567 & 0.274 & 0.411 & 0.252 & 0.188 \\
        StreamPETR \cite{wang2023exploring} & C & ResNet-50 & 256 × 704 & 54.0 & 43.2 & 0.581 & 0.272 & 0.413 & 0.295 & 0.195 \\
        SparseBEV \cite{liu2023sparsebev} & C & ResNet-50 & 256 × 704 & 54.5 & 43.2 & 0.606 & 0.274 & 0.387 & 0.251 & 0.186 \\
        CRN \cite{kim2023crn} & C+R & ResNet-50 & 256 × 704 & 56.0 & 49.0 & 0.487 & 0.277 & 0.542 & 0.344 & 0.197 \\
        RCBEVDet \cite{lin2024rcbevdet} & C+R & ResNet-50 & 256 × 704 & 56.8 & 45.3 & 0.486 & 0.285 & 0.404 & \textbf{0.220} & 0.192 \\
        CRT-Fusion \cite{kim2024crt} & C+R & ResNet-50 & 256 × 704 & 57.2 & 50.0 & 0.499 & 0.277 & 0.531 & 0.261 & 0.192 \\
        \rowcolor[gray]{0.9} IMKD (Ours) & C+R & ResNet-50 & 256 × 704 & \textbf{61.0} & \textbf{51.6} & \textbf{0.444} &\textbf{0.259} & \textbf{0.384} & 0.229 & \textbf{0.160} \\
        \midrule
        \midrule
        RICCARDO \cite{long2025riccardo} & C+R & ResNet101 & 1408 × 512 & 62.2 & \textbf{54.4} & 0.481 & 0.266 & \textbf{0.325} & 0.237 & 0.189 \\
        \rowcolor[gray]{0.9} IMKD (Ours) & C+R & ResNet101 & 1408 × 512 & \textbf{62.7} & 53.9 & \textbf{0.417} &\textbf{0.255} & 0.348 & \textbf{0.235} & \textbf{0.158} \\
        \bottomrule
    \end{tabular}
    \caption{Comparison of 3D object detection performance on the nuScenes~\cite{caesar2020nuscenes} validation set. ‘C’ and ‘R’ denote camera and radar, respectively. Methods utilizing future frames, test-time augmentation, and CBGS~\cite{zhu2019class} are excluded to ensure fairness. 
    The upper block reports comparisons restricted to BEVDepth with ResNet-50, while the lower block extends to ResNet-101 backbones and includes RICCARDO~\cite{long2025riccardo} for completeness.}
    \label{tab:comparison_val}
\end{table*}

\begin{table*}[b]
    \centering
    \setlength{\tabcolsep}{4pt} 
    \renewcommand{\arraystretch}{1.1}
    \begin{tabular}{l|c|c|c|c|c|c|c|c}
        \toprule
        Method & Input & NDS↑ & mAP↑ & mATE↓ & mASE↓ & mAOE↓ & mAVE↓ & mAAE↓ \\
        \midrule
        $\text{PGD}$ \cite{wang2022probabilistic} & C & 44.8 & 38.6 & 0.626 & 0.245 & 0.451 & 1.509 & 0.127 \\
        $\text{SparseBEV}$ \cite{liu2023sparsebev} & C & 67.5 & 60.3 & 0.425 & 0.239 & 0.311 & 0.172 & 0.116 \\
        MVFusion \cite{wu2023mvfusion} & C+R & 51.7 & 45.3 & 0.569 & 0.246 & 0.379 & 0.781 & 0.128 \\
        CRN \cite{kim2023crn} & C+R & 62.4 & 57.5 & 0.416 & 0.264 & 0.456 & 0.365 & 0.130 \\
        RCBEVDet \cite{lin2024rcbevdet} & C+R & 63.9 & 55.0 & 0.390 & \textbf{0.234} & 0.362 & 0.259 & 0.113 \\
        HyDRa \cite{wolters2025unleashing} & C+R & 64.2 & 57.4 & 0.398 & 0.251 & 0.423 & 0.249 & 0.122 \\
        HVDetFusion \cite{lei2307hvdetfusion} & C+R & 67.4 & 60.9 & 0.379 & 0.243 & 0.382 & 0.172 & 0.132 \\
        SparseBEV+RICCARDO \cite{long2025riccardo} & C+R &\textbf{69.5} & \textbf{63.0} & \textbf{0.363} & 0.240 & 0.311 & \textbf{0.167} & 0.118 \\
        \rowcolor[gray]{0.9} IMKD (Ours) & C+R & 67.0 & 61.0 & 0.401 & 0.249 & \textbf{0.305} & 0.238 & \textbf{0.102} \\
        \bottomrule
    \end{tabular}
    \caption{Comparison of 3D object detection performance on the nuScenes~\cite{caesar2020nuscenes} test set. ‘C’ and ‘R’ represent camera and radar, respectively.}
    \label{tab:comparison_test_all}
\end{table*}

To ensure a fair and meaningful comparison, we primarily benchmark IMKD against radar-camera fusion methods that share the same foundational settings. Specifically, we focus on approaches that adopt BEVDepth~\cite{li2023bevdepth} with a ResNet-50~\cite{he2016deep} backbone, avoiding discrepancies introduced by stronger visual encoders. We also exclude methods that leverage CBGS~\cite{zhu2019class}, test-time augmentation, or future frames, as such enhancements can distort the true impact of the fusion strategy. All comparisons are conducted on the nuScenes~\cite{caesar2020nuscenes} validation set, where the backbone architecture and image resolution are consistent across methods, unlike the test set, where configurations often vary. IMKD is the first distillation-driven framework to surpass the performance of standard radar-camera fusion methods, elevating knowledge distillation from a regularization tool to a core mechanism for advancing 3D detection performance.

As an exception, we additionally report results for RICCARDO~\cite{long2025riccardo}, which employs SparseBEV \cite{liu2023sparsebev} with a ResNet-101 \cite{he2016deep} backbone rather than BEVDepth \cite{li2023bevdepth} with ResNet-50 \cite{he2016deep}. While this setting is not strictly comparable to our fairness-controlled benchmark, it provides useful context on how IMKD scales with stronger visual encoders. To avoid misleading comparisons, we align RICCARDO’s \cite{long2025riccardo} results with our own ResNet-101 \cite{he2016deep} BEVDepth \cite{li2023bevdepth} variant, and present these separately in \cref{tab:comparison_val} under a distinct block. This highlights that IMKD maintains its advantage even when evaluated under higher-capacity camera backbones, demonstrating robustness across configurations.

These improvements stem from IMKD’s fusion-aware and signal-sensitive design. By incorporating intensity-aware distillation and fusion-based supervision, IMKD captures fine-grained signal reliability and cross-modal interactions that traditional fusion models overlook. As a result, IMKD not only bridges the gap between handcrafted fusion and learned fusion but also pushes the performance frontier for camera-radar 3D object detection.

We further report results on the nuScenes \cite{caesar2020nuscenes} test set to contextualize IMKD against the latest benchmark entries, as shown in \cref{tab:comparison_test_all}. While this comparison is not strictly fair, methods employ heterogeneous camera backbones (e.g., SparseBEV \cite{liu2023sparsebev} in RICCARDO \cite{long2025riccardo}) and varying image resolutions, it provides a broader view of IMKD’s standing. Despite these differences, IMKD achieves performance highly competitive with state-of-the-art methods, while remaining the only knowledge-distillation-based approach among the top-performing entries on the benchmark. This highlights both the practicality and the effectiveness of IMKD in advancing camera–radar 3D detection under challenging real-world settings.

\subsection{Comparison on VoD Dataset} \label{supp:vod_ds_comparison}

To evaluate the generalization of IMKD beyond the nuScenes \cite{caesar2020nuscenes} dataset, we conduct experiments on the View-of-Delft (VoD) \cite{apalffy2022} dataset, which provides synchronized LiDAR, camera, and 3+1D radar sensors, with the radar capturing elevation in addition to range, azimuth, and Doppler. This richer radar representation presents a more challenging detection scenario compared to the sparse 2+1D radar in nuScenes \cite{caesar2020nuscenes}.

As reported in \cref{tab:vod_comparison}, IMKD achieves strong performance across all categories, demonstrating competitive results relative to existing camera-radar methods. In particular, IMKD maintains high AP in both the entire annotated area and the region of interest, indicating that the intensity-guided distillation framework effectively transfers LiDAR knowledge and enhances fused representations even under different radar characteristics.

These results validate that our method generalizes robustly to other datasets and radar configurations, confirming that intensity-aware multi-level knowledge distillation can consistently improve cross-modal 3D detection beyond the original nuScenes \cite{caesar2020nuscenes} setting.

\begin{table*}[t]
    \centering
    \begin{tabular}{l| c | c c c |c | c c c| c}
    \hline
    \multirow{2}{*}{Method} & \multirow{2}{*}{Input} & \multicolumn{4}{c|}{AP in Entire Annotated Area (\%)} & \multicolumn{4}{c}{AP in Region of Interest (\%)} \\
     &  & Car & Pedestrian & Cyclist & mAP & Car & Pedestrian & Cyclist & mAP \\
    \hline
    PointPillars \cite{lang2019pointpillars} & R & 37.06 & 35.04 & 63.44 & 45.18 & 70.15 & 47.22 & 85.07 & 67.48 \\
    RadarPillarNet \cite{yang2020radarnet} & R & 39.30 & 35.10 & 63.63 & 46.01 & 71.65 & 42.80 & 83.14 & 65.86 \\
    RCFusion \cite{zheng2023rcfusion} & C+R & 41.70 & 38.95 & 68.31 & 49.65 & 71.87 & 47.50 & 88.33 & 69.23 \\
    RCBEVDet \cite{lin2024rcbevdet} & C+R & 40.63 & 38.86 & \textbf{70.48} & 49.99 & 72.48 & 49.89 & 87.01 & 69.80 \\
    IMKD (Ours) & C+R & \textbf{47.55} & \textbf{45.51} & 68.40 & \textbf{53.81} & \textbf{89.13} & \textbf{57.10} & \textbf{89.56} & \textbf{78.59} \\
    \hline
    \end{tabular}
    \caption{Comparison of 3D object detection results on the VoD \cite{apalffy2022} validation set. The region of interest is the driving corridor near the ego-vehicle. AP thresholds are set to 0.5 for cars, 0.25 for pedestrians, and 0.25 for cyclists.}
    \label{tab:vod_comparison}
\end{table*}

\subsection{BEV Segmentation} \label{supp:bev_segmentation}




Our method leverages knowledge distillation from LiDAR and label guidance to enhance camera-radar features, enabling precise segmentation of road elements such as drivable areas, lanes, and crossings. LiDAR distillation refines spatial accuracy, improving object boundaries and structural details. We use mean Intersection over Union (mIoU) as the primary metric, following \cite{philion2020lift}. As shown in \cref{tab:bev_segmentation}, our approach achieves an mIoU of 62.2, demonstrating effective segmentation with real-time performance.

\begin{table}[H]
    \centering
    \renewcommand{\arraystretch}{1} 
    \setlength{\tabcolsep}{1.0pt} 
    \begin{tabular}{l|c|c|c|c|c}
        \toprule
        Method & Input & Backbone & mIoU↑ & Veh↑ & D.A↑ \\
        \midrule
        BEVFormer-S \cite{li2203bevformer} & C & R101 & 48.4 & 43.2 & 80.7 \\
        CRN \cite{kim2023crn} & C+R & R50 & - & 58.8 & \textbf{82.1} \\
        Simple-BEV++$^\dagger$ \cite{schramm2024bevcar} & C+R & R101 & 55.4 & 52.7 & 77.7 \\
        BEVGuide \cite{man2023bev} & C+R & EffNet & 60.0 & 59.2 & 76.7 \\
        BEVCar \cite{schramm2024bevcar} & C+R & R101 & 61.0 & 57.3 & 81.8 \\
        \rowcolor[gray]{0.9} IMKD (Ours) & C+R & R101 & \textbf{62.2} & \textbf{60.5} & 81.9 \\
        \bottomrule
    \end{tabular}
    \caption{Comparison of BEV semantic segmentation on the nuScenes \cite{caesar2020nuscenes} validation set. ‘C’ and ‘R’ represent camera and radar, respectively. ‘D.A’ denotes drivable area. $\dagger$ indicates a Simple-BEV \cite{harley2023simple} model customized by BEVCar \cite{schramm2024bevcar}.}
    \vspace{-0.5cm}
    \label{tab:bev_segmentation}
\end{table}

\begin{table*}[b]
    \centering
    \renewcommand{\arraystretch}{1}
    \setlength{\tabcolsep}{1pt}
    \begin{tabular}{l|c|c| c| c| c| c| c| c| c| c| c| c}
        \toprule
        Method & Input & Car & Truck & Bus & Trailer & C.V. & Ped. & M.C. & Bicycle & T.C. & Barrier & mAP \\
        \midrule
        CenterFusion \cite{nabati2021centerfusion} & C+R & 52.4 & 26.5 & 36.2 & 15.4 & 5.5 & 38.9 & 30.5 & 22.9 & 56.3 & 47.0 & 33.2 \\
        CRAFT \cite{kim2023craft} & C+R & 69.6 & 37.6 & 47.3 & 20.1 & 10.7 & 46.2 & 39.5 & 31.0 & 57.1 & 51.1 & 41.1 \\
        CRN \cite{kim2023crn} & C+R & 71.9 & 42.4 & 51.1 & 27.1 & 16.2 & 46.6 & 54.0 & 44.2 & 56.7 & 61.6 & 47.1 \\
        \textbf{IMKD (Ours)} & C+R & \textbf{75.3}\textsuperscript{4.7\%} & \textbf{50.9}\textsuperscript{20.0\%} & \textbf{55.6}\textsuperscript{8.8\%} & \textbf{28.6}\textsuperscript{5.5\%} & \textbf{20.6}\textsuperscript{27.2\%} & \textbf{55.1}\textsuperscript{18.2\%} & \textbf{54.5}\textsuperscript{0.9\%} & \textbf{51.1}\textsuperscript{15.6\%} & \textbf{62.2}\textsuperscript{9.7\%} & \textbf{62.1}\textsuperscript{0.8\%} & \textbf{51.6}\textsuperscript{9.6\%} \\
        \bottomrule
    \end{tabular}
    \caption{Per-class comparisons on the nuScenes \cite{caesar2020nuscenes} validation set. ‘C.V.’, ‘Ped.’, ‘M.C.’, and ‘T.C.’ denote construction vehicle, pedestrian, motorcycle, and traffic cone, respectively. All results are sourced from MMDetection3D and official implementations, except CRN, which was reproduced using its official GitHub repository.}
    \label{tab:per_class}
\end{table*}

\begin{table*}[t]
    \centering
    \small
    \renewcommand{\arraystretch}{1.4}
    \setlength{\tabcolsep}{0.8pt}
    \begin{tabular}{l|c|c| c| c| c| c| c| c| c| c| c| c}
        \toprule
        Method & Input & Car & Truck & Bus & Trailer & C.V. & Ped. & M.C. & Bicycle & T.C. & Barrier & mAP \\
        \midrule
        CenterNet \cite{zhou2019objects} & C & 48.4 & 23.1 & 34.0 & 13.1 & 3.5 & 37.7 & 24.9 & 23.4 & 55.0 & 45.6 & 30.6 \\
        CenterFusion \cite{nabati2021centerfusion} & C+R & 52.4\textsuperscript{8.3\%} & 26.5\textsuperscript{14.7\%} & 36.2\textsuperscript{6.5\%} & 15.4\textsuperscript{17.5\%} & 5.5\textsuperscript{57.1\%} & 38.9\textsuperscript{3.2\%} & 30.5\textsuperscript{22.5\%} & 22.9\textsuperscript{-1.4\%} & 56.3\textsuperscript{2.4\%} & 47.0\textsuperscript{3.0\%} & 33.2\textsuperscript{0.6\%} \\
        \midrule
        CRAFT-I \cite{kim2023craft} & C & 52.4 & 25.7 & 30.0 & 15.8 & 5.4 & 39.3 & 28.6 & 29.8 & 57.5 & 47.8 & 33.2 \\
        CRAFT \cite{kim2023craft} & C+R & 69.6\textsuperscript{32.8\%} & 37.6\textsuperscript{46.3\%} & 47.3\textsuperscript{57.6\%} & 20.1\textsuperscript{27.2\%} & 10.7\textsuperscript{98.1\%} & 46.2\textsuperscript{17.5\%} & 39.5\textsuperscript{38.1\%} & 31.0\textsuperscript{4.0\%} & 57.1\textsuperscript{-0.7\%} & 51.1\textsuperscript{7.0\%} & 41.1\textsuperscript{23.8\%} \\
        \midrule
        BEVDepth \cite{li2023bevdepth} & C & 55.3 & 25.2 & 37.8 & 16.3 & 7.6 & 36.1 & 31.9 & 28.6 & 53.6 & 55.9 & 34.8 \\
        CRN \cite{kim2023crn} & C+R & 71.9\textsuperscript{30.0\%} & 42.4\textsuperscript{67.9\%} & 51.1\textsuperscript{35.2\%} & 27.1\textsuperscript{66.9\%} & 16.2\textsuperscript{113.2\%} & 46.6\textsuperscript{29.1\%} & 54.0\textsuperscript{69.2\%} & 44.2\textsuperscript{54.2\%} & 56.7\textsuperscript{5.8\%} & 61.6\textsuperscript{10.2\%} & 47.1\textsuperscript{35.6\%} \\
        \midrule
        BEVDepth \cite{li2023bevdepth} & C & 55.3 & 25.2 & 37.8 & 16.3 & 7.6 & 36.1 & 31.9 & 28.6 & 53.6 & 55.9 & 34.8 \\
        IMKD (Ours) & C+R & 75.3\textsuperscript{36.6\%} & 50.9\textsuperscript{101.2\%} & 55.6\textsuperscript{57.3\%} & 28.6\textsuperscript{75.2\%} & 20.6\textsuperscript{171.1\%} & 55.1\textsuperscript{52.4\%} & 54.5\textsuperscript{71.8\%} & 51.1\textsuperscript{78.7\%} & 62.2\textsuperscript{10.5\%} & 62.1\textsuperscript{9.6\%} & 51.6\textsuperscript{47.6\%} \\
        \bottomrule
    \end{tabular}
    \caption{Per-class comparisons on the nuScenes \cite{caesar2020nuscenes} validation set, evaluating each camera + radar network against its corresponding camera-only variant. ‘C.V.’, ‘Ped.’, ‘M.C.’, and ‘T.C.’ denote construction vehicle, pedestrian, motorcycle, and traffic cone, respectively. All results are sourced from MMDetection3D and official implementations, except CRN, which was reproduced using its official GitHub repository.}
    \label{tab:per_class_individual}
\end{table*}

\subsection{Per-Class Performance Analysis}
In \cref{tab:per_class}, we compare per-class performance across different camera-radar fusion methods, using a fixed resolution of 256×704 and the ResNet-50 backbone for consistency.

In \cref{tab:per_class_individual}, we compare each camera-only network with its camera+radar variant on the nuScenes \cite{caesar2020nuscenes} validation set. The results show that radar significantly improves performance in most classes. Using the same camera-only baseline as CRN, our method outperforms previous approaches in several categories.

Our IMKD method consistently achieves the highest mAP, with notable improvements in Truck, Bus, C.V., Pedestrian, and Bicycle. This demonstrates the effectiveness of our fusion strategy in handling various object types, particularly for smaller or more dynamic objects where radar data can be especially beneficial. The improvements in classes like Pedestrian and Bicycle, where radar information is typically sparse, further validate the robustness of our approach.

Key to this performance is our knowledge distillation framework, which refines the fusion of camera and radar features through LiDAR-guided and label-based distillation, ensuring that radar signals contribute meaningfully to object detection rather than introducing noise. This structured supervision enhances detection accuracy, leading to more reliable and consistent object localization across all categories.

Overall, our results show that distilling knowledge into the fused modality improves camera-radar fusion, significantly boosting performance.

\section{Qualitative Analysis} \label{supp:Qualitative}

\begin{figure*}[t]
  \centering
    \includegraphics[width=1\linewidth]{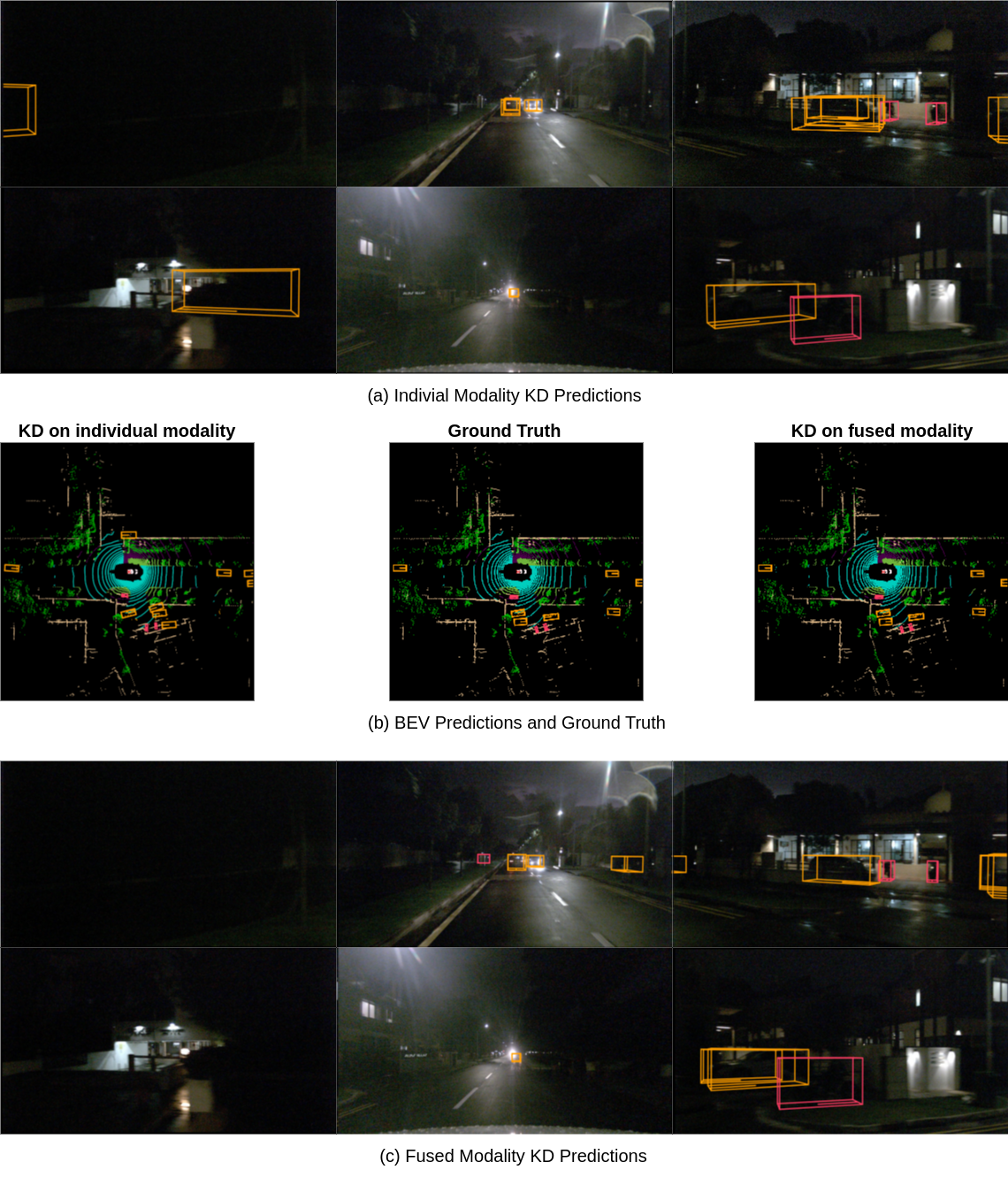}
  \caption{Qualitative results of our proposed IMKD method on night scenes from the nuScenes~\cite{caesar2020nuscenes} dataset. (a) shows camera-view predictions from individual modality distillation baselines. (b) presents BEV predictions: left shows individual modality predictions, middle is the ground truth, and right shows IMKD results. (c) displays IMKD’s predictions across six camera views, illustrating improved detection quality under challenging low-light conditions.}
  \label{fig:qualitative_night}
\end{figure*}

\begin{figure*}[t]
  \centering
    \includegraphics[width=1\linewidth]{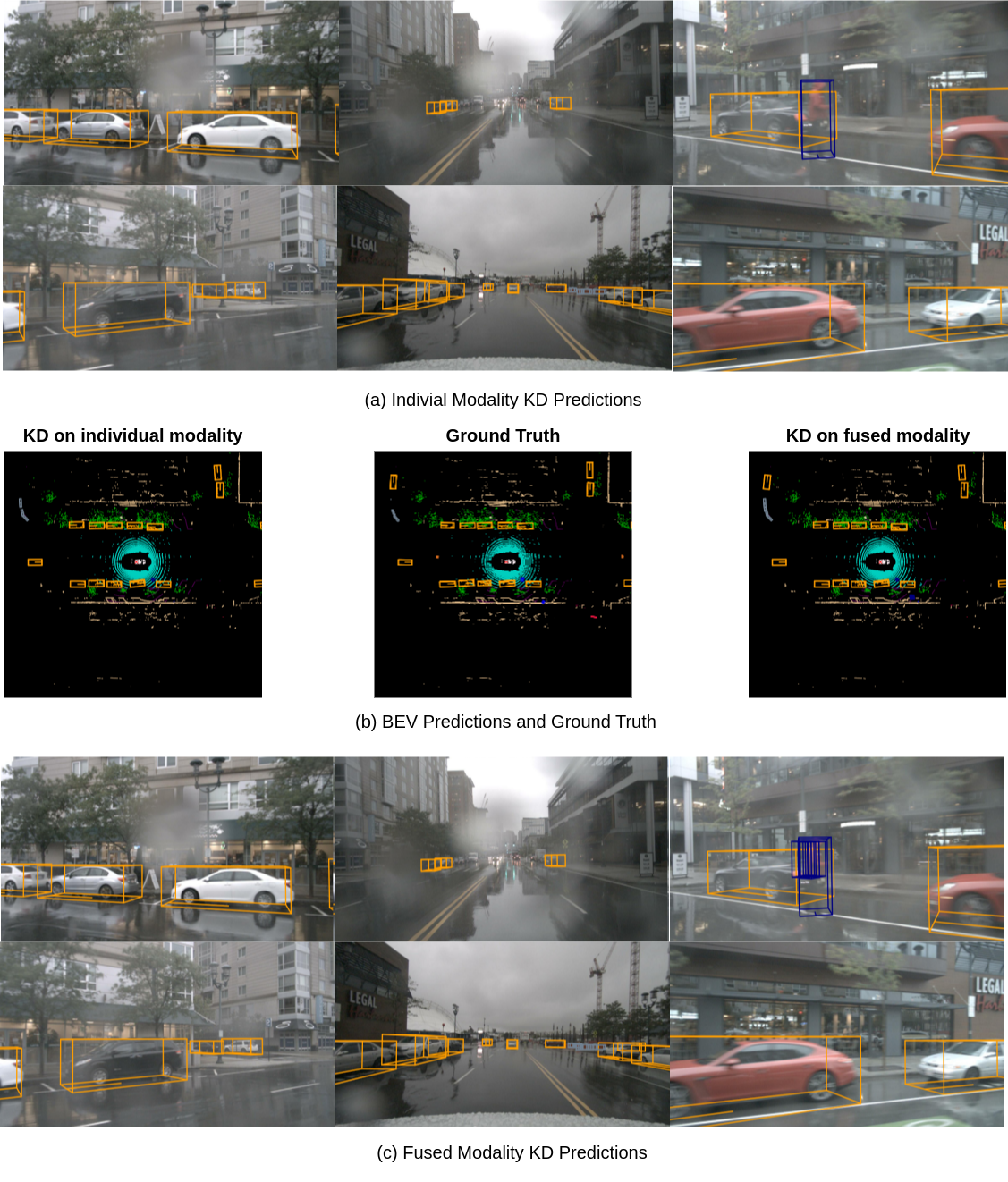}
  \caption{Qualitative results of our proposed IMKD method on rainy scenes from the nuScenes \cite{caesar2020nuscenes} dataset. (a) shows camera-view predictions from individual modality distillation baselines. (b) presents BEV predictions: left shows individual modality predictions, middle is the ground truth, and right shows IMKD results. (c) displays IMKD’s predictions across six camera views, illustrating improved detection quality under challenging low-light conditions.}
  \label{fig:qualitative_rain}
\end{figure*}

\begin{figure*}[t]
  \centering
    \includegraphics[width=1\linewidth]{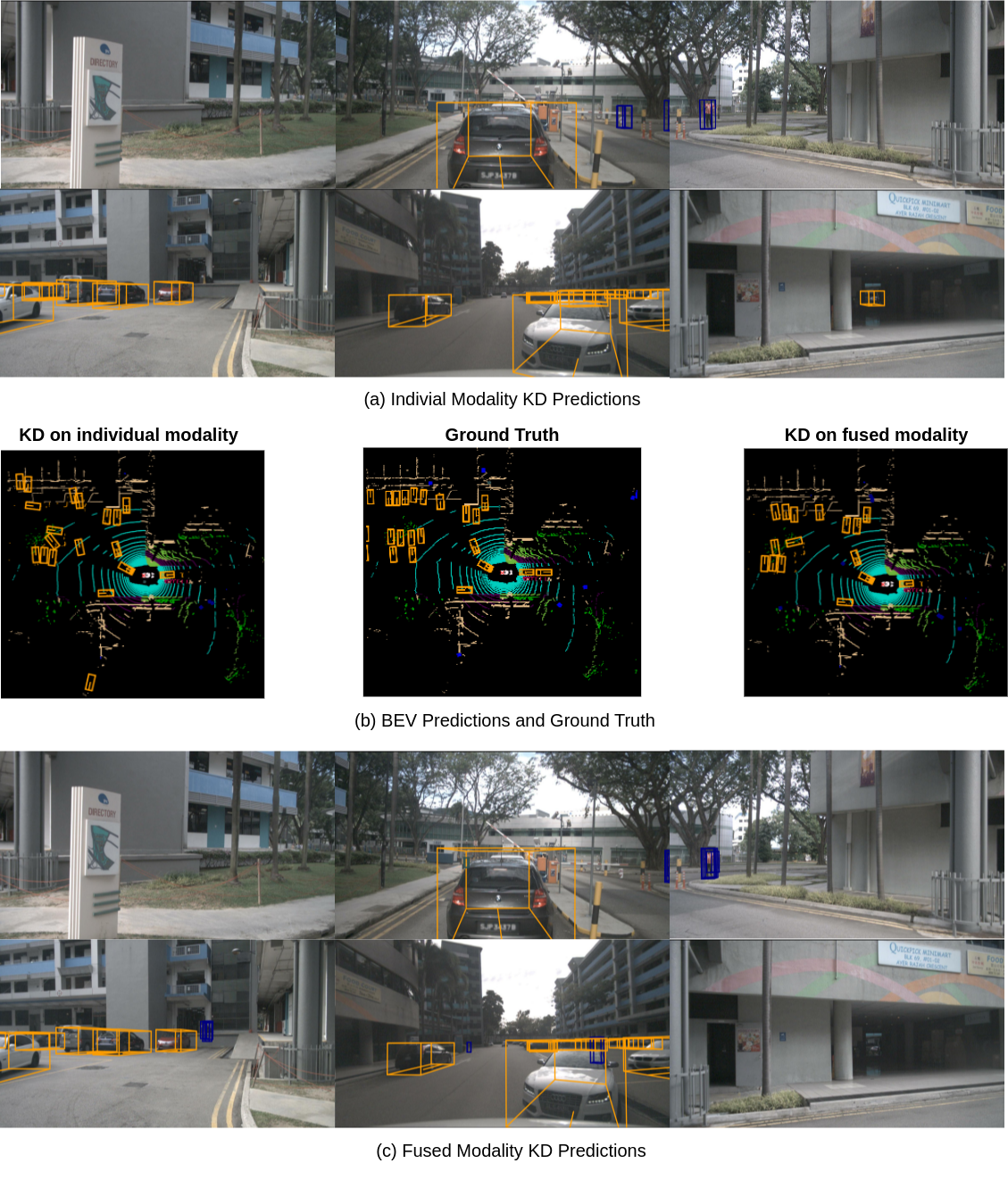}
  \caption{Qualitative results of our proposed IMKD method on day scenes from the nuScenes \cite{caesar2020nuscenes} dataset. (a) shows camera-view predictions from individual modality distillation baselines. (b) presents BEV predictions: left shows individual modality predictions, middle is the ground truth, and right shows IMKD results. (c) displays IMKD’s predictions across six camera views, illustrating improved detection quality under challenging low-light conditions.}
  \label{fig:qualitative_day}
\end{figure*}

We present additional qualitative results under varying weather and lighting conditions, including rainy, nighttime, and daytime scenarios, from the nuScenes~\cite{caesar2020nuscenes} dataset. As shown in \cref{fig:qualitative_day,fig:qualitative_rain,fig:qualitative_night}, IMKD consistently performs better than individual modality distillation baselines, particularly under challenging scenarios like rain and low light.

In these adverse conditions, conventional single-modality distillation models often fail to detect occluded or distant objects. In contrast, IMKD consistently performs better by utilizing intensity-guided fusion and merged-modality knowledge distillation. The fusion mechanism dynamically weighs radar and camera features based on signal confidence, while the distillation strategy transfers depth and structural cues from LiDAR into the joint camera-radar representation. This enables IMKD to produce more accurate and robust object detections, boxes with better translation, orientation, and scale accuracy than baselines, crucial under low visibility where conventional methods struggle to infer reliable geometry. These improvements are clearly reflected in both BEV and multi-view camera predictions.

